\documentclass[10pt,twocolumn,letterpaper]{article}

\usepackage[pagenumbers]{cvpr} %

\usepackage{algorithm}
\usepackage{algpseudocode}

\usepackage{amssymb}
\usepackage{amsthm}

\newtheorem{definition}{Definition}[section]
\newtheorem{lemma}[definition]{Lemma}

\newtheorem{proposition}[definition]{Proposition}

\newtheorem{remark}[definition]{Remark}

\definecolor{cvprblue}{rgb}{0.21,0.49,0.74}
\usepackage[pagebackref,breaklinks,colorlinks,allcolors=cvprblue]{hyperref}

\hypersetup{
  pdftitle={DraftKV: Protecting Persistent Worlds through Hierarchical Streaming Retrieval},
  pdfauthor={Xiaoling Zhou, Haopeng Li, Yufa Zhou, Rui Ding, Yu Li, Jun Lin, Enze Xie, Chunhua Shen, Xuan Shen},
  pdfsubject={Historical KV retrieval for video world models},
  pdfkeywords={DraftKV, video world models, KV cache, streaming attention},
  hypertexnames=false
}

\def\confName{CVPR}
\def\confYear{2027}

\title{
DraftKV: Protecting Persistent Worlds through\\ Hierarchical Streaming Retrieval
}

\newcommand{\affil}[1]{%
  {\fontsize{9}{11}\selectfont #1}%
}

\author{
Xiaoling Zhou$^{1,*}$ \quad
Haopeng Li$^{2,*}$ \quad
Yufa Zhou$^{3}$ \quad
Rui Ding$^{4}$ \quad
Yu Li$^{1}$ \\
Jun Lin$^{4}$ \quad
Enze Xie$^{5}$ \quad
Chunhua Shen$^{1,\dagger}$ \quad
Xuan Shen$^{1,\dagger}$ \\[0.7em]
\affil{
$^{1}$Zhejiang University \quad
$^{2}$The Hong Kong University of Science and Technology (Guangzhou)
}\\[0.2em]
\affil{
$^{3}$Duke University \quad
$^{4}$Nanjing University \quad
$^{5}$NVIDIA Research
}\\[0.5em]
{\small
$^{*}$Equal contribution.
\quad
$^{\dagger}$Corresponding authors.
}\\[0.4em]
{\tt\small endlesstilly@gmail.com}\\[0.5em]
{\small
\textbf{Code:}
\href{https://github.com/TillyEndless/DraftKV}
{\texttt{github.com/TillyEndless/DraftKV}}
}
}

\begin{document}
\maketitle

\begin{abstract}

Interactive video world models enable action- and camera-controlled exploration, yet struggle to maintain consistent spatial geometry and appearance when revisiting previously explored locations.
Historical KV caches help auto-regressive video diffusion models maintain long-term consistency, but their retrieval and attention computation introduce substantial latency as history grows.
We propose \textbf{DraftKV}, a training-free framework that retrieves historical KV caches based on geometric grounding and query-conditioned relevance, while combining CPU offloading with tile-wise streaming attention to reduce GPU memory usage and inference latency.
Theoretically, we show that draft relevance reproduces the mean branch of full-resolution block relevance exactly, so its deviation is confined to the $\beta$-weighted max branch, and we bound retrieval utility loss in terms of the measured Top-K selection overlap.
We further show that draft relevance cannot distinguish blocks with similar pooled descriptors regardless of camera pose, motivating geometric grounding before refinement.
Then, we introduce streaming CPU-to-GPU Attention, which overlaps CPU-to-GPU KV transfers with attention computation through tile-wise streaming, hiding transfer latency while maintaining bounded GPU memory usage.
Experiments across diverse model families show that DraftKV achieves superior generation quality and consistency 
while achieving \textbf{3.27$\times$} end-to-end speedup and \textbf{42.2\%} peak GPU memory reduction over other CPU-offloaded KV cache retrieval method.
Notably, DraftKV enables high-quality interactive world generation with historical KV retrieval on a single H100 GPU using SolarWM-33B.

\end{abstract}
\section{Introduction}
\label{sec:introduction}

\begin{figure}[t]
   \centering
   \includegraphics[width=\columnwidth]{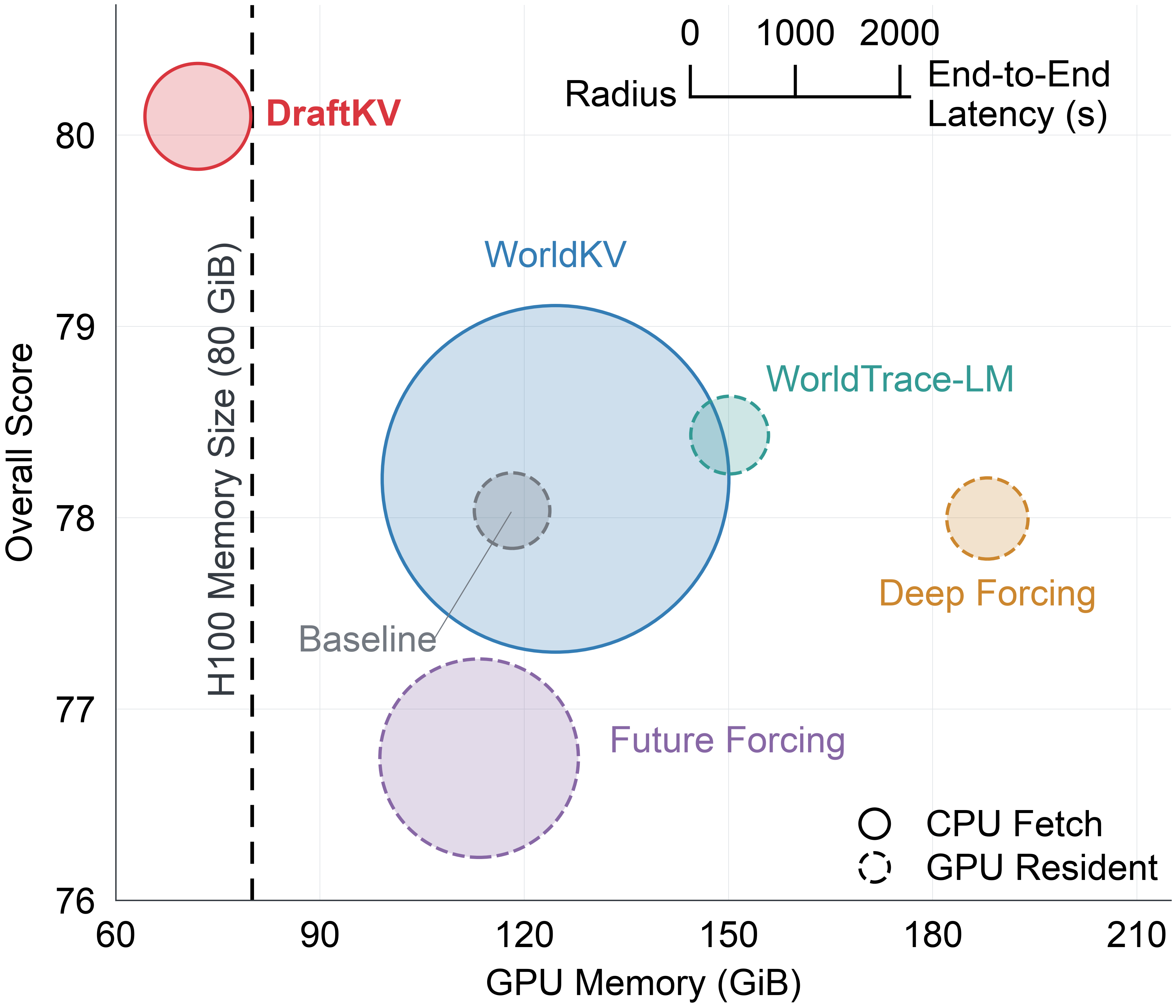}
   \vspace{-6mm}
   \caption{
    Comparison of long-horizon KV retrieval methods on SolarWM-33B.
    Note that the vertical dashed line marks the 80\,GiB memory capacity of a single H100 GPU. 
    }
   \label{fig:comparision}
   \vspace{-4mm}
\end{figure}

Interactive video world models enable open-ended exploration through action and camera control, yet preserving a persistent world over long horizons remains challenging~\cite{matrixgame2,lingbotworld}.
This limitation manifests as a failure to recover previously observed spatial structure and appearance upon revisiting the same location and viewpoint.
Historical key--value (KV) caches serve as a memory medium for autoregressive video diffusion transformers~\cite{worldkv}.
Thus, preserving useful historical KV for future revisits may help improve long-horizon world persistence.

However, retaining all historical KV-cache to avoid premature forgetting quickly becomes prohibitive in terms of both storage and full-history attention costs.
Sliding-window and KV compression methods reduce GPU memory usage but may discard historical information essential for future scene revisits~\cite{longlive,worldtrace}.
In contrast, we observe that selectively retrieving relevant historical KV can preserve generation quality without retaining the full history.
These observations raise two key questions: how to identify and retrieve the necessary historical KV, and how to store and fetch historical KV efficiently.

Camera pose provides spatial context for identifying previously visited viewpoints~\cite{worldkv,contextasmemory}.
However, pose-based retrieval may introduce irrelevant KV due to coarse granularity and potential camera trajectory drift.
Retrieving irrelevant historical KV wastes GPU memory and incurs unnecessary data transfers.
In contrast, query--key affinity captures fine-grained visual relevance but may retrieve similar-looking regions from incorrect locations without spatial constraints.
These complementary properties motivate a hierarchical retrieval strategy combining camera-based spatial grounding with query-guided fine-grained selection.

Offloading historical KV-cache to CPU reduces GPU memory consumption~\cite{infinigen,shadowkv}, but introduces H2D transfer latency and GPU memory overhead when loading retrieved KV for active-layer attention. To address these challenges, we leverage the tile-wise computation of FlashAttention~\cite{flashattention} to stream retrieved KV from CPU to GPU. By overlapping data transfers with attention computation and reusing GPU buffers, we reduce transfer latency while keeping active KV memory bounded regardless of retrieval size.

We propose \textbf{DraftKV}, a training-free framework for memory-efficient historical KV retrieval in large-scale autoregressive video world models.
DraftKV employs a coarse-to-fine retrieval strategy: \textbf{Draft Grounding} first identifies spatially relevant historical KV using camera geometry, while \textbf{Draft Refinement} then selects query-relevant KV through low-resolution draft attention.
Then, we empirically observe that our coarse-to-fine retrieval identifies visually similar regions across historical and current frames. 
We then show that draft relevance reproduces the mean branch of full-resolution block relevance exactly, so that its deviation is confined to the $\beta$-weighted max branch. 
Separately, we bound the retrieval utility lost by drafting in terms of the measured agreement between draft and full-resolution KV selection. 
Finally, we show that draft relevance cannot separate blocks whose pooled descriptors are close, whatever their camera poses, while an idealized frustum-overlap rule discards no co-visible history in a static scene, which together motivate geometric grounding before refinement.
Apart from retrieval accuracy, GPU memory consumption remains a major challenge, as retaining excessive historical KV or loading all retrieved KV at once can easily exhaust GPU memory. 
While our coarse-to-fine retrieval reduces redundant KV selection, DraftKV further offloads full-resolution historical KV to CPU memory, retaining only compact draft representations on the GPU.
To further reduce peak GPU memory usage, we introduce \textbf{Streaming CPU-to-GPU Attention (SCG-Attention)}, which processes retrieved KV in tiles, overlapping CPU-to-GPU transfers with attention computation and recycling GPU buffers after each tile is processed. This design hides transfer latency while bounding GPU-resident retrieved KV to a fixed number of tiles, regardless of history length.
Evaluations spanning diverse autoregressive video world models and long-horizon benchmarks show that DraftKV improves long-range visual persistence while simultaneously reducing GPU memory requirement. 
DraftKV achieves better long-horizon generation persistence and overall performance compared to other existing methods, while reducing much more peak GPU memory requirement.
Notably, DraftKV enables historical KV retrieval for SolarWM-33B video world models on a single 80 GiB GPU.

\begin{figure}[t]
    \centering
    \includegraphics[width=\columnwidth]{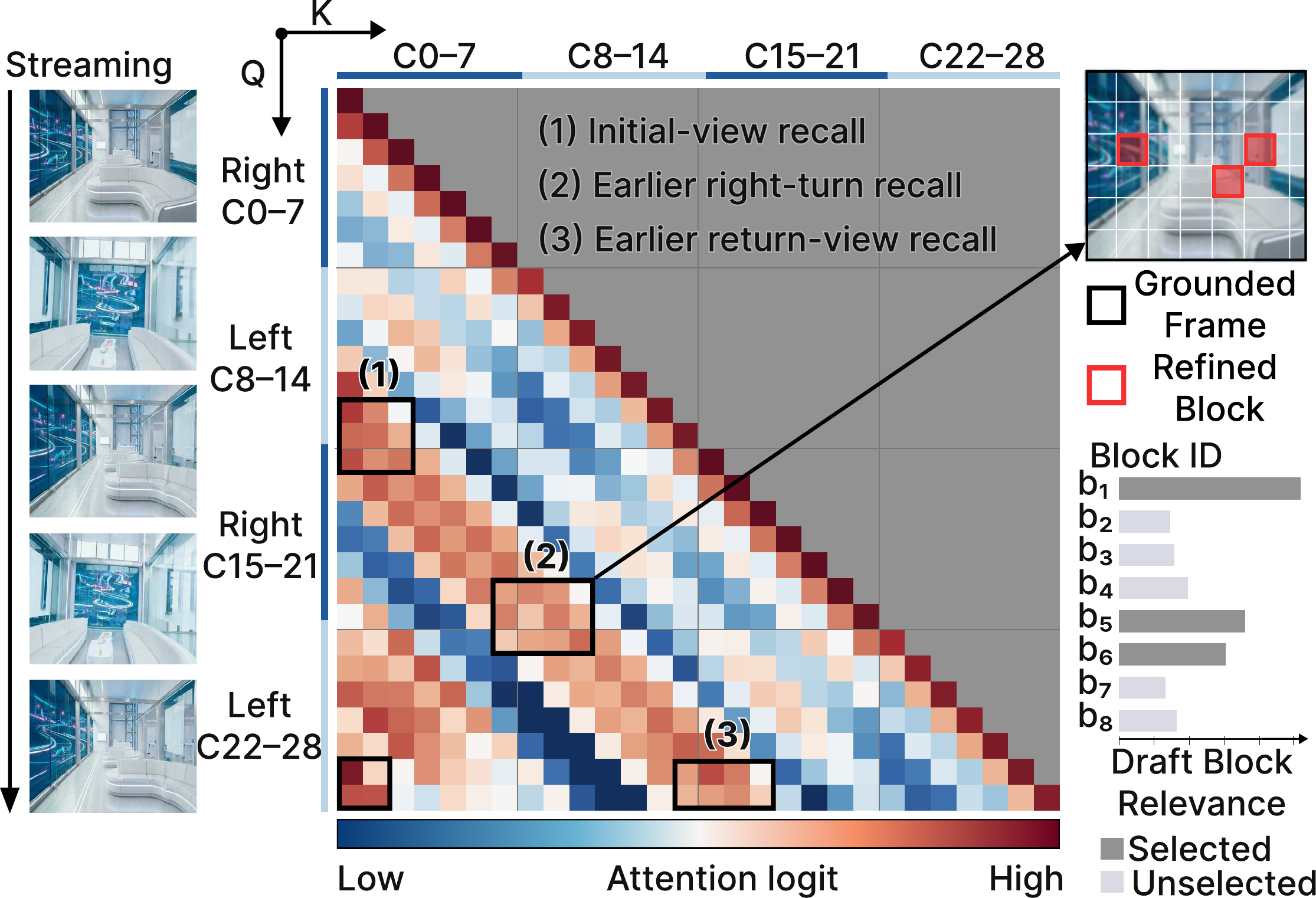}
    \vspace{-6mm}
    \caption{
    \textbf{Illustration of Hierarchical Draft Retrieval.}
    The attention-logit map highlights three historical revisits (1--3) across different camera viewpoints. Draft Grounding identifies spatially corresponding frames, while Draft Refinement selects query-relevant KV blocks within them, as illustrated on the right.
    }
    \label{fig:retrieval}
    \vspace{-4mm}
\end{figure}

\begin{figure*}[!t]
    \centering
    \includegraphics[width=\textwidth]{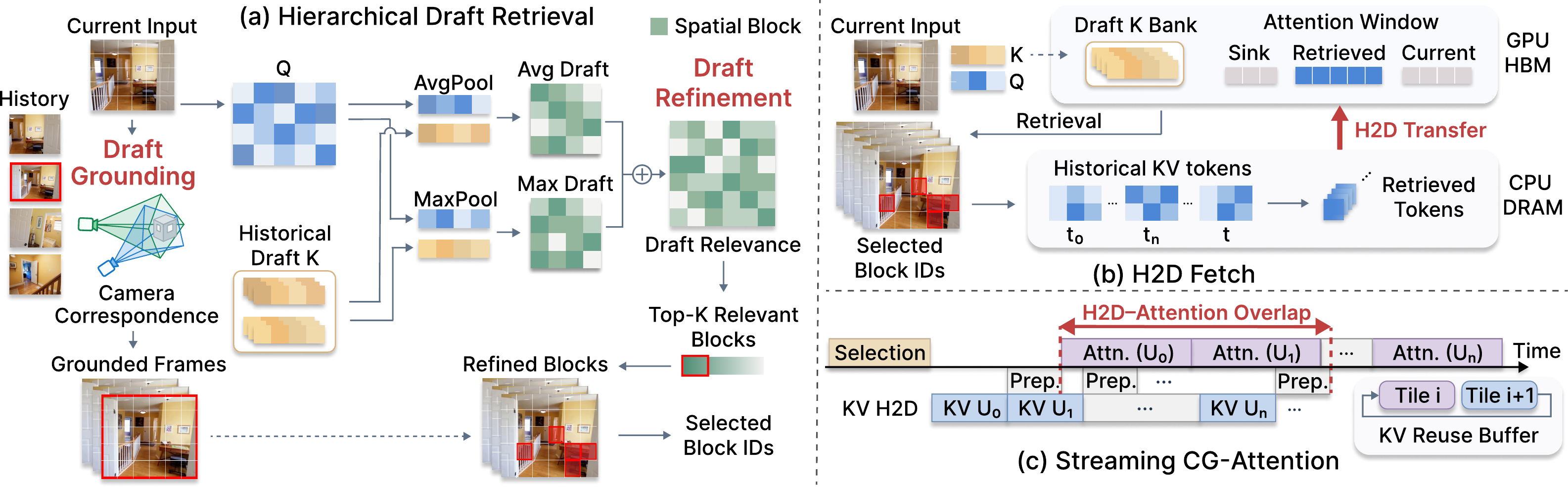}
    \vspace{-6mm}
    \caption{
    \textbf{Overview of DraftKV.}
    (a) Hierarchical Draft Retrieval first grounds historical observations through camera correspondence, then refines them at block granularity using draft block relevance.
    (b) The compact Draft K Bank remains GPU-resident for retrieval, while full-resolution historical KV is stored in CPU memory and fetched on demand.
    (c) SCG-Attention overlaps tile-wise H2D transfer with attention computation through a bounded KV Reuse Buffer, limiting active GPU residency.
    }
    \label{fig:framework}
    \vspace{-4mm}
\end{figure*}

Our main contributions are summarized as follows:
\begin{itemize}
\item \textbf{Hierarchical Draft Retrieval.} We introduce a coarse-to-fine retrieval scheme that combines Draft Grounding for spatially consistent historical KV retrieval with Draft Refinement for fine-grained query-relevant KV selection. Meanwhile, we establish an exact relation between draft and full-resolution block relevance, a retrieval utility bound in terms of the measured Top-$K$ overlap, and an aliasing bound and co-visibility guarantee that together motivate geometric grounding.

\item \textbf{Streaming CPU-to-GPU Attention.} We develop a tile-wise CPU-to-GPU attention pipeline that hides H2D latency behind computation and bounds active KV residency independently of retrieved-history length.

\item \textbf{Empirical Validation.} We quantitatively demonstrate that DraftKV improves long-horizon generation persistence by \textbf{13.7\%} and overall performance by \textbf{5.0\%} over native inference, 
while achieving a \textbf{3.27$\times$} end-to-end speedup and \textbf{42.2\%} peak GPU memory reduction over other CPU-offloaded KV cache retrieval method.
\end{itemize}

\section{Related Work}
\label{sec:related_work}
\noindent\textbf{Long-Horizon Memory in Video World Models.}
Long-horizon video generation mainly relies on bounded local contexts~\citep{self_forcing,longlive,rolling_forcing}, compressed or recurrent memory~\citep{worldtrace,hlawm,relic}, or explicit historical retrieval~\citep{longlive_rag,worldkv,omnimem,worldattention}.
The first two bound memory by discarding or compressing distant observations, whereas retrieval preserves explicit history and selectively reactivates it when needed.
DraftKV follows this retrieval paradigm, focusing on accurate historical KV selection and efficient access.

\noindent\textbf{Historical KV Retrieval.}
Existing methods retrieve history using different signals and granularities.
LongLive-RAG~\citep{longlive_rag} retrieves historical latents through learned embeddings; WorldKV~\citep{worldkv} uses camera and action correspondence for chunk-level KV retrieval; and GEAR~\citep{gear} establishes geometry-based token correspondences.
Query-guided methods~\citep{infinigen,shadowkv,omnimem} instead estimate KV relevance from attention.
DraftAttention~\citep{draftattention,draftattention2} further shows that low-resolution query--key interactions can approximate fine-grained attention relevance.
Different from these approaches, DraftKV combines camera-based spatial grounding with draft query--key refinement, enabling training-free fine-grained retrieval of original historical KV without full-resolution scoring over the entire history.

\noindent\textbf{Offloaded KV Access and Attention Execution.}
CPU offloading reduces persistent GPU memory but introduces H2D transfer and active-KV materialization overhead.
InfiniGen~\citep{infinigen} uses predictive prefetching, ShadowKV~\citep{shadowkv} reduces transfer through compact representations, WorldPager~\citep{worldpager} employs pose-guided prefetching across GPU/CPU/NVMe, and WorldAttention~\citep{worldattention} combines hierarchical KV management with sparse attention and optimized kernels.
In contrast, DraftKV streams selected KV during attention execution: SCG-Attention overlaps tile-wise H2D transfer with computation and reuses bounded GPU buffers, avoiding full selected-set materialization before attention.

\begin{figure*}[!t]
    \centering
    \begin{subfigure}[!t]{0.24\textwidth}
        \centering
        \includegraphics[width=\linewidth]{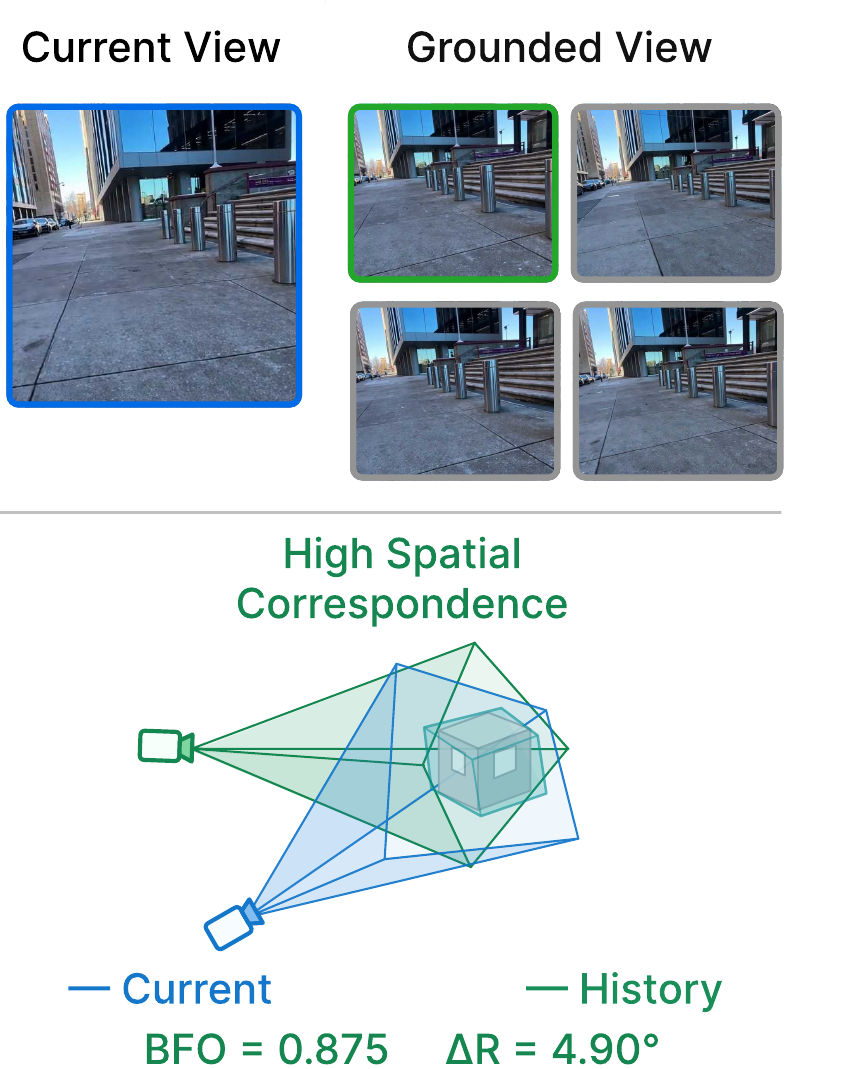}
        \caption{Draft Grounding}
        \label{fig:motivation_grounding}
    \end{subfigure}
    \hfill
    \begin{subfigure}[!t]{0.24\textwidth}
        \centering
        \includegraphics[width=\linewidth]{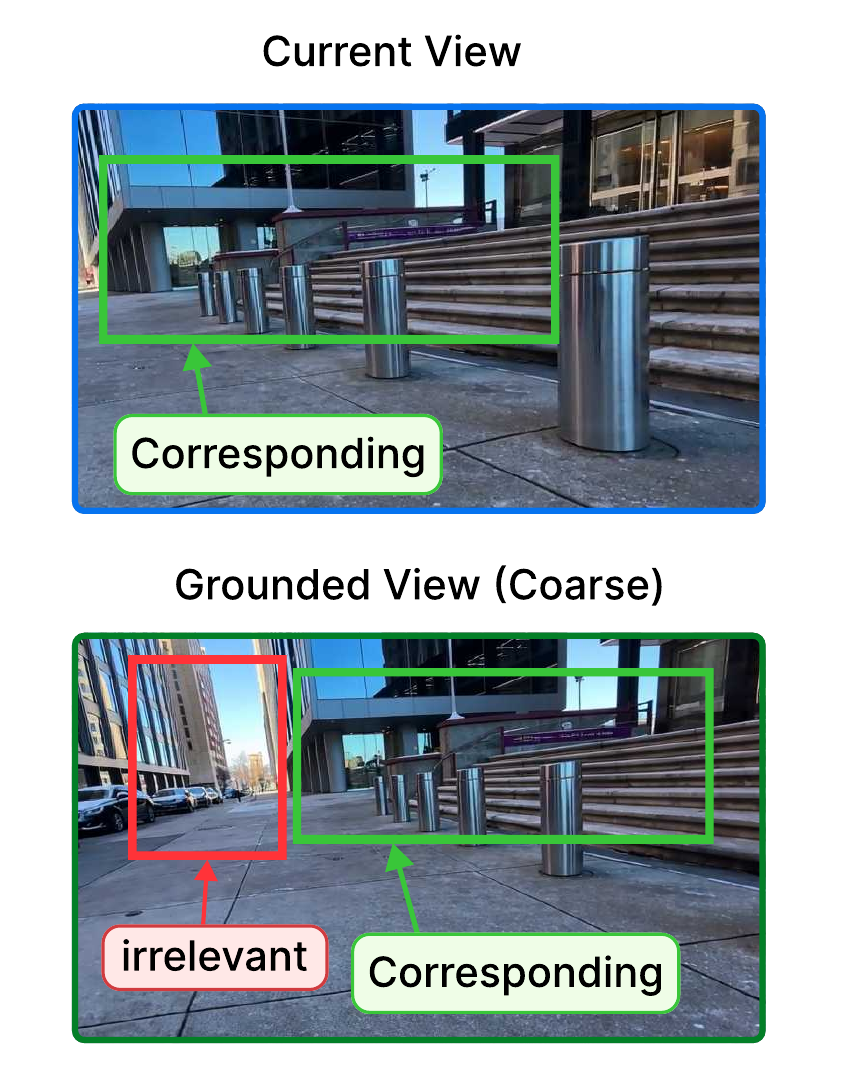}
        \caption{Grounding is Locally Coarse}
        \label{fig:motivation_partial}
    \end{subfigure}
    \hfill
    \begin{subfigure}[!t]{0.24\textwidth}
        \centering
        \includegraphics[width=\linewidth]{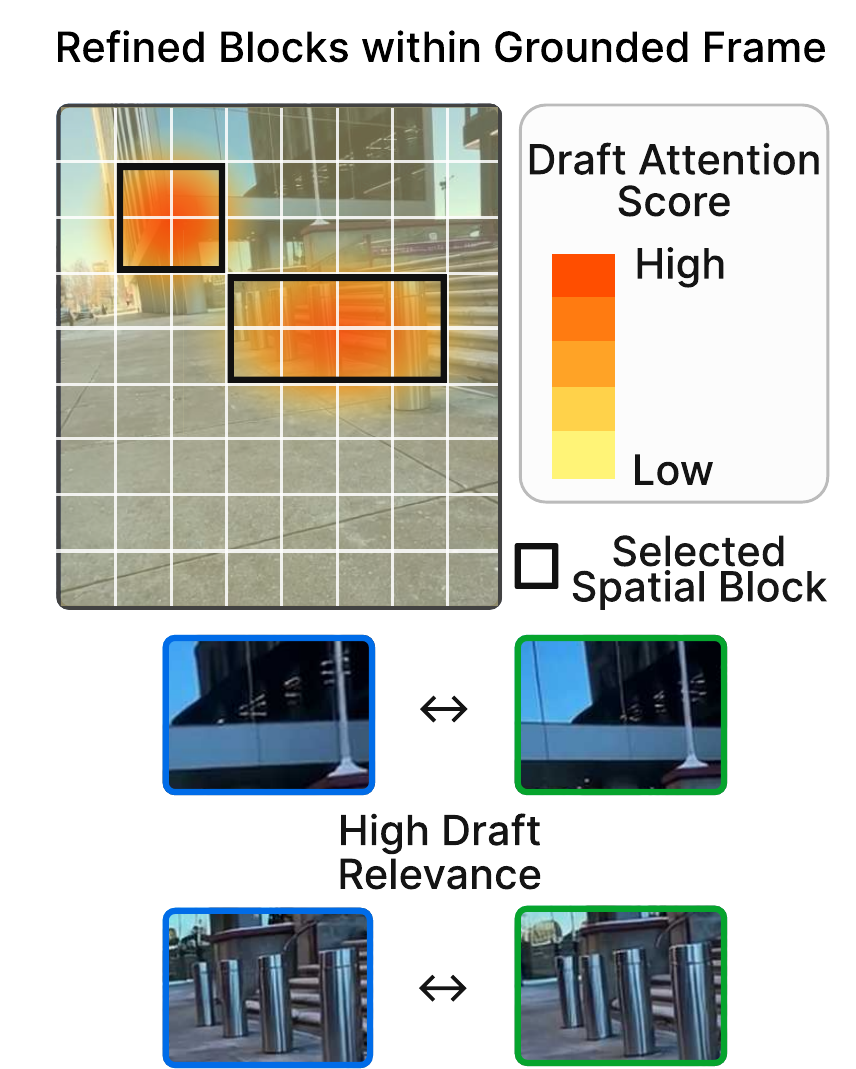}
        \caption{Draft Refinement}
        \label{fig:motivation_refinement}
    \end{subfigure}
    \hfill
    \begin{subfigure}[!t]{0.24\textwidth}
        \centering
        \includegraphics[width=\linewidth]{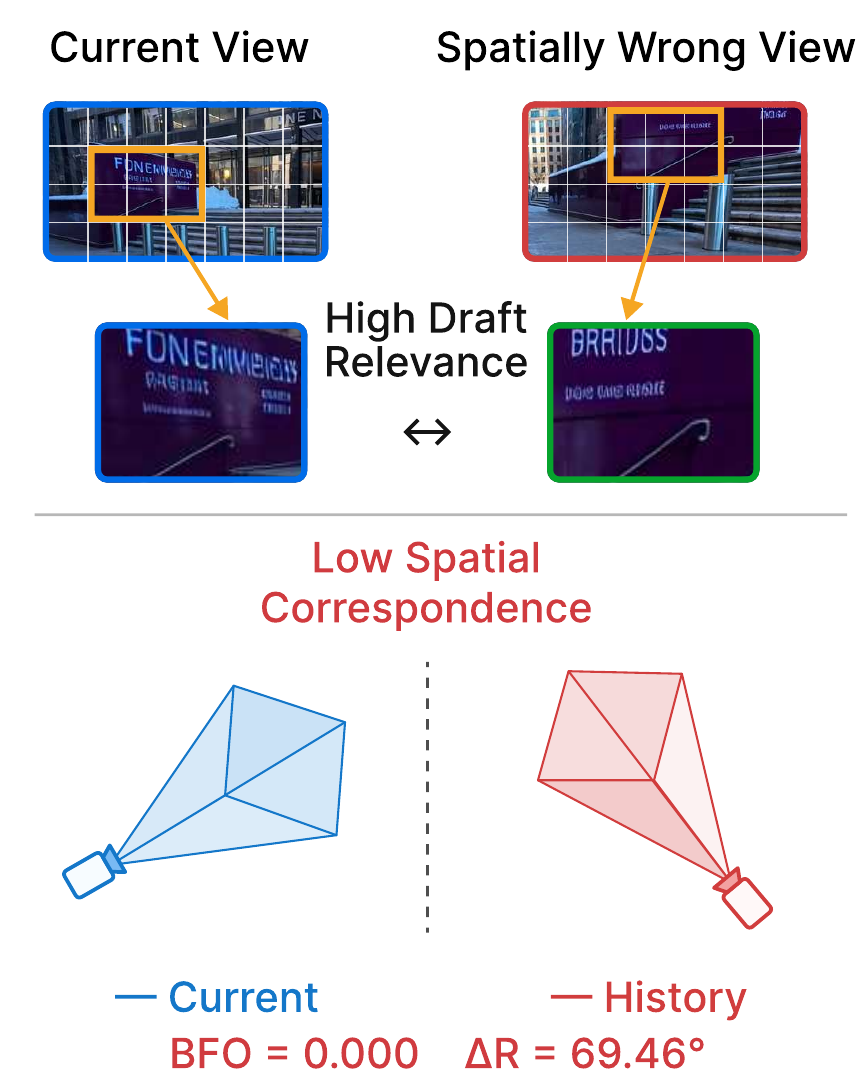}
        \caption{Draft-Only Spatial Ambiguity}
        \label{fig:motivation_ambiguity}
    \end{subfigure}
    \vspace{-2mm}
    \caption{
    \textbf{Why Hierarchical Draft Retrieval?}
    Draft Grounding identifies spatially corresponding historical views, but observation-level grounding remains locally coarse under partial visual overlap.
    Draft Refinement selects query-relevant blocks within grounded views, while draft relevance alone lacks explicit world-space provenance and may favor spatially mismatched history.
    Hierarchical Draft Retrieval combines both signals for spatially grounded and fine-grained historical retrieval.
    }
    \label{fig:retrieval_motivation}
    \vspace{-4mm}
\end{figure*}

\section{Methodology}

\subsection{Correlation Investigation}
\label{sec:correlation-investigation}

Historical retrieval for world models has to meet two requirements at once, spatial consistency and fine-grained appearance fidelity.
First, camera geometry provides spatial correspondence for historical retrieval, as illustrated in Fig.~\ref{fig:motivation_grounding}.
We then use camera geometry to localize candidate history, which we term \textbf{Draft Grounding}:
\begin{equation}
g_t(i)=S_{\mathrm{BFO}}(\mathcal{F}_t,\mathcal{F}_i),
\qquad
\mathcal{C}_t=
\operatorname*{TopM}_{i\in\mathcal{H}_t} g_t(i),
\label{eq:draft_grounding}
\end{equation}
where $\mathcal{H}_t$ is the eligible history, the past chunks still available for retrieval at step $t$, and $S_{\mathrm{BFO}}$ is a bilateral frustum-overlap score, a sampled proxy for co-visibility between the current and historical views. Grounding ranks whole chunks, so $\mathcal{C}_t$ retains the top-$M$ chunks with all of their frames; Appendix~\ref{sec:appendix-theory} states how this realizes the geometric criterion and where it departs from it.

\begin{figure}
    \centering
    \vspace{-4mm}
    \includegraphics[width=\linewidth]{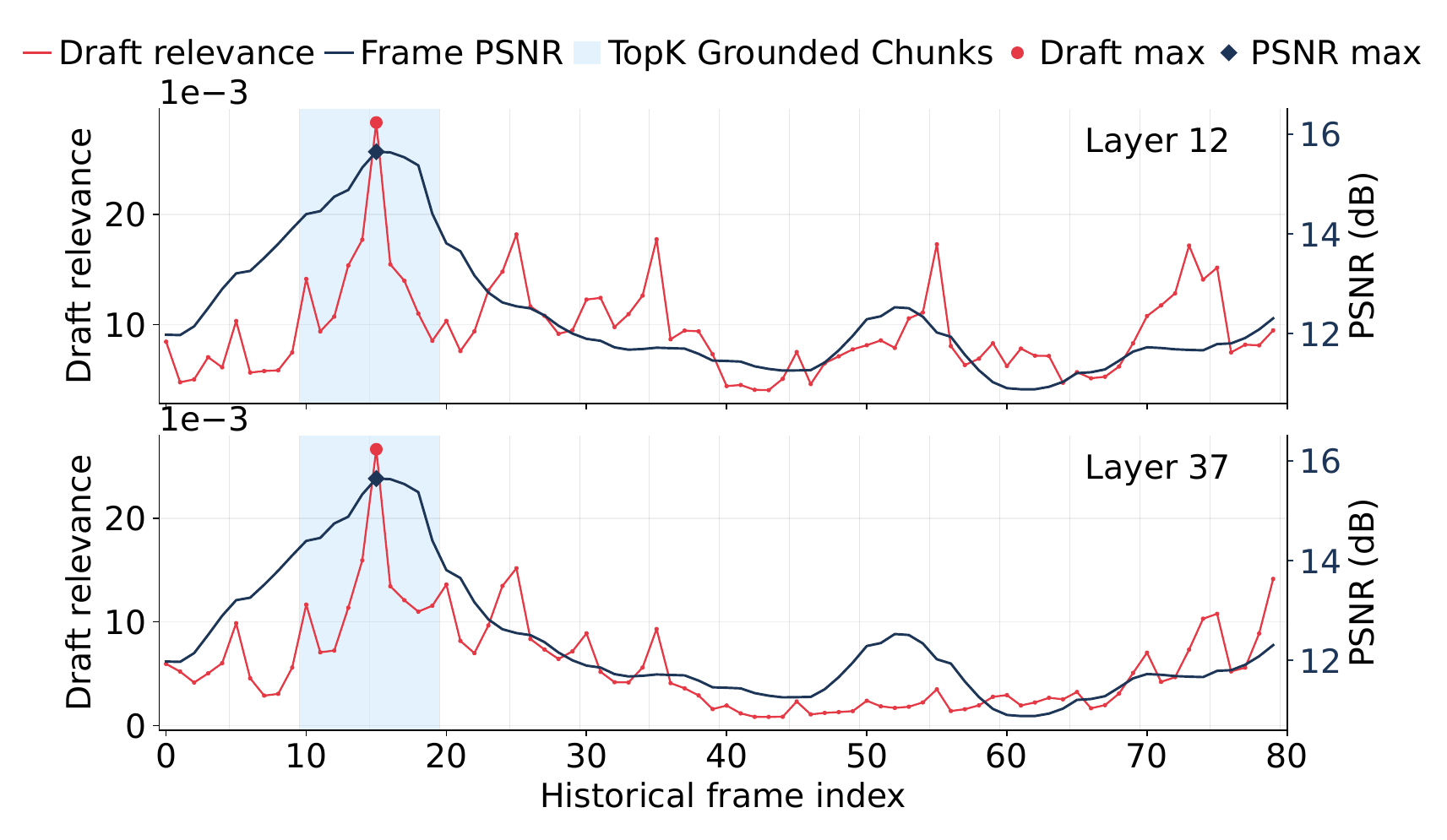}
    \vspace{-8mm}
    \caption{PSNR measures similarity to the current revisit frame. Across layers, Draft Refinement aligns with the PSNR peak within the grounded range, enabling accurate revisit-frame localization.}
    \label{fig:draft_psnr}
    \vspace{-4mm}
\end{figure}

However, grounded chunks may include irrelevant history, as visualized in Fig.~\ref{fig:motivation_partial}.
As shown in Fig.~\ref{fig:draft_psnr}, Draft Grounding successfully identifies the most relevant historical frame but also retains irrelevant content, introducing unnecessary GPU memory overhead. This motivates fine-grained refinement within the grounded candidates.

Draft attention score provides a low-cost estimate of fine-grained visual relevance.
We therefore design Draft Refinement, which uses block-level draft attention scores to select history within grounded frames, as shown in Fig.~\ref{fig:motivation_refinement}.
We define the draft attention score of spatial block $b$ as
\begin{equation}
\tilde r_t(b)
=
\operatorname{DraftScore}(Q_t,\widetilde K_b),
\label{eq:draft_importance}
\end{equation}
where $Q_t$ denotes the current queries, pooled per spatial query region, and
$\widetilde K_b=(\bar{k}_b,k_b^{\max})$
is the compact dual-pooled key descriptor stored for block $b$. Scores are
normalized over the candidate blocks and the protected recent frames that are always retained.
(see Sec.~\ref{sec:theoretical-analysis} for details)

Draft Refinement then selects the top-$K$ high-score blocks among the grounded candidates, with the protected recent frames kept regardless:
\begin{equation}
\mathcal{R}_t
=
\operatorname*{TopK}_{b\in\mathcal{B}(\mathcal{C}_t)}
\tilde r_t(b),
\label{eq:draft_refinement}
\end{equation}
where $\mathcal{B}(\mathcal{C}_t)$ denotes all historical KV blocks contained in the grounded chunks $\mathcal{C}_t$ selected by Draft Grounding.

\subsection{Theoretical Analysis}
\label{sec:theoretical-analysis}

Section~\ref{sec:correlation-investigation} motivates draft block relevance as a low-cost proxy for full-resolution block relevance. We formalize both scores and show the draft score exactly recovers the mean component of full-resolution relevance. Since KV similarity does not guarantee spatial correspondence, we restrict ranking to geometrically grounded candidates.

\paragraph{Setup.}
Fix a generated chunk $t$, writing $t$ also for its current frame where
frame-level geometry is involved. Both scores are normalized over the scoring
domain $\mathcal{B}_t=\mathcal{B}(\mathcal{C}_t\cup\mathcal{P}_t)$, with
$\mathcal{C}_t$ the observations selected by Draft Grounding and
$\mathcal{P}_t$ the protected recent frames; the latter participate in
normalization but not in competitive retrieval, whose candidate domain is
$\mathcal{B}^{\mathrm{ret}}_t=\mathcal{B}(\mathcal{C}_t)$. Index heads by
$h=1,\dots,H$ and query regions by $r=1,\dots,N_q$, let $\mathcal{Q}_r$ and
$\mathcal{K}_b$ be their token sets, and let
$\ell^{(h)}_{ij}=\langle q_i,k_j\rangle/\sqrt{d}$ be the token-pair logit on the
post-QKNorm, post-RoPE representations the model itself consumes; head
superscripts on $q$ and $k$ are suppressed where unambiguous.

\begin{definition}[Block Relevance]
\label{def:importance}
Let $\mu^{(h)}_{r,b}$ and $X^{(h)}_{r,b}$ be the mean and the maximum of
$\ell^{(h)}_{ij}$ over $\mathcal{Q}_r\times\mathcal{K}_b$, and
$P^{(h),\mathrm{mean}}_{r,b}$, $P^{(h),\max}_{r,b}$ their softmax
normalizations over $\mathcal{B}_t$. The full-resolution block relevance is
\begin{equation}
  I_b=\frac{1}{HN_q}\sum_{h,r}
    \Big[(1-\beta)\,P^{(h),\mathrm{mean}}_{r,b}+\beta\,P^{(h),\max}_{r,b}\Big].
  \label{eq:importance}
\end{equation}
Draft block relevance $\widehat{I}_b$ is defined identically with
$\mu^{(h)}_{r,b}$ and $X^{(h)}_{r,b}$ replaced by
$\widehat{\mu}^{(h)}_{r,b}=\langle\bar q_r,\bar k_b\rangle/\sqrt{d}$ and
$\widehat{X}^{(h)}_{r,b}=\langle q^{\max}_r,k^{\max}_b\rangle/\sqrt{d}$, the token averages and
channel-wise maxima held in the Draft K Bank. It instantiates
$\mathrm{DraftScore}$ in Eq.~(\ref{eq:draft_importance}), so
$\tilde r_t(b)=\widehat{I}_b$.
\end{definition}

\begin{lemma}[Dual-Pooling Relation]
\label{lem:pooling}
$\widehat{\mu}^{(h)}_{r,b}=\mu^{(h)}_{r,b}$ exactly, for every head and every
pair of regions. Consequently the mean branch of Eq.~\eqref{eq:importance} is
identical under the two scores and
$\big|\widehat{I}_b-I_b\big|\le\beta$ for every $b$.
\end{lemma}

The mean branch carries the remaining weight $1-\beta=0.96$ and is reproduced
without error, so the entire approximation is confined to the max branch,
consistent with the near-unit rank correlation in
Appendix~\ref{sec:appendix-retrieval}. 

Retrieval depends only on the induced ordering of relevance scores, and where the two orderings disagree the cost is bounded by how far
they disagree.

\begin{proposition}[Utility Transfer]
\label{prop:utility}
Let $S,\widehat{S}$ be the size-$K$ selections over
$\mathcal{B}^{\mathrm{ret}}_t$ under full-resolution and draft relevance, and
let $\rho=|S\cap\widehat{S}|/K$. For any block-level utility
$u:\mathcal{B}^{\mathrm{ret}}_t\!\to\![u_{\min},u_{\max}]$,
\begin{equation}
  \frac{1}{K}\sum_{b\in\widehat{S}}u_b
  \ \ge\
  \frac{1}{K}\sum_{b\in S}u_b
  \;-\;(1-\rho)\,(u_{\max}-u_{\min}).
  \label{eq:utility-transfer}
\end{equation}
\end{proposition}

The selections differ in $K(1-\rho)$ elements, each contributing at most $u_{\max}-u_{\min}$ to the utility loss. The bound holds for every such $u$, including a perceptual
similarity the draft score never observes, so with the Draft Refinement
Recall@$K$ of Appendix~\ref{sec:appendix-retrieval}, the mean utility forfeited by
drafting is at most $1-\rho=4.38\%$ of the utility range, whichever utility is
chosen. The figure is a coefficient on that range rather than a measured drop
in any particular metric, and $\rho$ is measured rather than bounded.

What that ordering ranks is content affinity alone. Draft relevance depends on
a historical block only through its pooled descriptors, so two blocks of $\mathcal{B}_t$ whose
pooled descriptors agree within $\epsilon$ in $\ell_2$, both descriptors and every
head, satisfy
$e^{-\eta}\le\widehat{I}_b/\widehat{I}_{b'}\le e^{\eta}$, where
$\eta=\epsilon\max_{h,r}\max\{\|\bar q^{(h)}_r\|_2,\|q^{(h),\max}_r\|_2\}/\sqrt{d}$,
whatever the camera poses of their source frames (derivation in
Appendix~\ref{sec:appendix-theory}). Pose enters the score only
through the keys and nothing makes it separate them, so a block retrieved on
relevance may come from anywhere that happens to look similar: the perceptual
aliasing of place recognition~\cite{cummins2008fabmap,lowry2016vpr}.
Restricting competitive retrieval to $\mathcal{B}(\mathcal{C}_t)$ supplies the
missing world-space constraint, and that constraint is a statement about the
scene rather than about the network.

\begin{proposition}[Co-visibility Necessity]
\label{prop:covisibility}
Assume a static scene, and assume frames $s$ and $t$ realize their commanded
camera poses, so that the view frusta $\mathcal{F}_s$ and $\mathcal{F}_t$
describe the content the model actually generates. If a scene point is observed
both by a historical block $b$ of frame $s$ and by the current query region
$r$, it lies in $\mathcal{F}_s\cap\mathcal{F}_t$, so that intersection is
non-empty. Consequently a rule retaining every frame whose frustum meets
$\mathcal{F}_t$ discards no block that observes a scene point also observed by
$r$.
\end{proposition}

The idealized rule therefore discards no co-visible evidence; Draft Grounding
approximates it under a fixed budget, so the guarantee belongs to the rule
rather than to its realization, with the approximations set out in
Appendix~\ref{sec:appendix-theory}. Frustum overlap uses no scene depth and so
resolves neither occlusion nor which regions within a retained frame correspond
to $r$. That residual is left to content relevance in Draft Refinement, which
is why the two signals compose in this order rather than acting independently.
\begin{remark}[Spatial granularity]
\label{rem:granularity}
Relevance is additive over a frame's blocks, $I_f=\sum_{b\in f}I_b$, so at a fixed
budget of $K$ blocks any whole-frame selection is itself a feasible block-level
selection and block-level Top-$K$ retains at least as much relevance, strictly
more unless the budget is exactly filled by frames none of whose blocks is
outranked by a block outside. Granularity pays when relevant blocks are spread over more frames
than the budget can hold whole.
\end{remark}

Furthermore, we provide detailed proofs of Lemma~\ref{lem:pooling} and Proposition~\ref{prop:covisibility} in Appendix~\ref{sec:appendix-theory}.

\begin{algorithm}[!t]
\caption{SCG-Attention for one layer}
\label{alg:streaming-cg}
\small
\begin{algorithmic}[1]
\Require queries $\{Q_g\}_{g=1}^{H/G}$; historical KV in CPU memory;
sink/current KV; KV Reuse Buffer slots $\mathcal{B}_{0:R-1}$
\Ensure attention outputs $\{O_g\}_{g=1}^{H/G}$

\State Partition each head group's KV window into disjoint
tiles $\{U_{g,j}\}_{j=1}^{M}$ of $\le C$ tokens, indexed in transfer order

\For{$k=0,\ldots,MH/G-1$} \Comment{head-group-major}
  \State $g\gets\lfloor k/M\rfloor+1$;
  $j\gets(k\bmod M)+1$; $r\gets k\bmod R$
  \State \textbf{if} $k\ge R$: wait until $\mathcal{B}_r$ is released
  \Comment{transfer}
  \State Fetch historical KV of $U_{g,j}$ into $\mathcal{B}_r$ if needed
  \State Wait for fetch; insert assigned sink/current KV
  \Comment{prepare}
  \State Apply temporal K-RoPE at assigned positions
  \State Wait for preparation \Comment{attend}
  \State $(O_{g,j},\lambda_{g,j})\gets
  \textsc{FlashAttn}(Q_g,U_{g,j})$
  \State Store $(O_{g,j},\lambda_{g,j})$;
  release $\mathcal{B}_r$
\EndFor
\State Merge stored partials into $\{O_g\}$ by
Eq.~\eqref{eq:tile-merge}
\end{algorithmic}
\end{algorithm}

\subsection{Streaming CPU-to-GPU Attention}
\label{sec:streaming-cg}

Hierarchical Draft Retrieval determines which historical KV each layer attends to; SCG-Attention determines how this KV reaches attention. Only the compact Draft K Bank stays on the GPU for retrieval, while full-resolution historical KV resides in pinned CPU memory and is fetched once a layer has selected its blocks (Fig.~\ref{fig:framework}(b)). Because the selection depends on the current query, this fetch lies on the critical path of every layer. Conventional execution further imposes \emph{selected-set readiness}: attention starts only after the entire retrieved set has been transferred to the GPU, which exposes the full H2D latency and requires a reuse workspace that grows with the retrieved context. 
The Draft K Bank stores descriptors under historical RoPE for retrieval, while full-resolution keys retain raw temporal components and receive re-indexed K-RoPE at their assigned attention positions.

\begin{table*}[!t]
\centering
\caption{Comparison of different methods on VBench-Long~\cite{vbenchlong}.
We extended VBench-Long to 60 seconds following LongLive~\cite{longlive}. 
}
\label{tab:vbench_results}
\vspace{-3mm}
\setlength{\tabcolsep}{4pt}
\renewcommand{\arraystretch}{1.1}
\small

\resizebox{\textwidth}{!}{%
\begin{tabular}{lccccccc|c|ccc}
\toprule

& \multicolumn{8}{c}{\textbf{Evaluation Scores}}
& \multicolumn{3}{c}{\textbf{Efficiency}} \\

\cmidrule(lr){2-9}
\cmidrule(lr){10-12}

\textbf{Method} &
\textbf{Subj.} $\uparrow$ &
\textbf{Bkg.} $\uparrow$ &
\textbf{Motion} $\uparrow$ &
\textbf{Dyn.} $\uparrow$ &
\textbf{Aesth.} $\uparrow$ &
\textbf{Imag.} $\uparrow$ &
\textbf{Flick.} $\uparrow$ &
\textbf{Quality} $\uparrow$ &
\textbf{E2E (s)} $\downarrow$ &
\textbf{FPS} $\uparrow$ &
\textbf{Mem. (GiB)} $\downarrow$ \\

\midrule

Baseline
& 96.96 & 95.82 & 98.92 & 38.79
& 59.65 & 71.99 & 97.97 & 82.07
& 331.88 & 2.88 & 24.64 \\

ShadowKV
& 97.67 & 96.11 & 99.20 & 17.04
& 58.51 & 70.63 & 98.66 & 80.64
& 1364.14 & 0.70 & 33.23 \\

FreeKV
& 98.78 & 97.28 & 99.39 & 6.49
& 60.26 & 70.63 & 99.09 & 80.82
& 5795.78 & 0.17 & 24.64 \\

InfiniGen
& 97.59 & 96.52 & 99.20 & 14.06
& 60.32 & 72.32 & 98.79 & 81.07
& 345.83 & 2.77 & 25.14 \\

InfLLM
& 97.06 & 95.89 & 99.05 & 35.81
& 58.53 & 71.39 & 98.28 & 81.81
& 910.23 & 1.05 & 24.64 \\

\cmidrule(lr){1-12}

Deep Forcing
& 96.54 & 95.70 & 98.67 & 50.21
& 57.75 & 71.34 & 97.34 & 82.07
& 200.25 & 4.78 & 26.64 \\

Future Forcing
& 95.44 & 95.11 & 98.66 & 63.67
& 53.67 & 68.65 & 97.70 & 81.88
& 372.73 & 2.57 & 27.15 \\

WorldTrace-LM
& 96.62 & 95.72 & 98.84 & 44.31
& 57.70 & 71.53 & 97.84 & 81.95
& 194.21 & 4.93 & 24.98 \\

\midrule

\textbf{DraftKV}
& 96.95 & 95.95 & 98.86 & 42.42
& 58.57 & 72.03 & 97.71 & \textbf{82.08}
& 242.87 & 3.94 & \textbf{24.64} \\

\bottomrule
\end{tabular}%
}
\vspace{-2mm}
\end{table*}

\begin{table*}[!t]
\centering
\caption{
Long-horizon world-model evaluation on ORBIA~\cite{orbia}.
WorldKV and DraftKV are CPU-offloaded retrieval methods. 
}
\label{tab:main_results}
\vspace{-3mm}
\setlength{\tabcolsep}{4pt}
\renewcommand{\arraystretch}{1.1}
\small

\resizebox{\textwidth}{!}{%
\begin{tabular}{llcccccc|c|ccc}
\toprule

& &
\multicolumn{7}{c}{\textbf{Evaluation Scores}} &
\multicolumn{3}{c}{\textbf{Efficiency}} \\

\cmidrule(lr){3-9}
\cmidrule(lr){10-12}

\textbf{Model} &
\textbf{Method} &
\textbf{Ctrl.} &
\textbf{Qual.}$\uparrow$ &
\textbf{InPres.}$\uparrow$ &
\textbf{Persist.}$\uparrow$ &
\textbf{3D Cons.}$\uparrow$ &
\textbf{Stab.}$\uparrow$ &
\textbf{Overall}$\uparrow$ &
\textbf{E2E (s)}$\downarrow$ &
\textbf{FPS}$\uparrow$ &
\textbf{Mem. (GiB)}$\downarrow$ \\

\midrule

\multirow{7}{*}{\makecell{LingBot\\1.3B}}
& Baseline
& \underline{49.38}
& \underline{73.92}
& 61.65
& \underline{56.67}
& \underline{58.42}
& \underline{78.08}
& \underline{64.11}
& 61.45 & 7.81 & 22.78\enspace(1.00$\times$) \\

& Deep Forcing
& 48.91
& 73.62
& \underline{61.85}
& 55.36
& 57.73
& 77.62
& 63.65
& 67.32 & 7.13 & 27.39\enspace(0.83$\times$) \\

& Future Forcing
& 45.92
& 73.38
& 61.41
& 54.48
& 55.32
& 76.04
& 62.15
& 165.31 & 2.90 & 22.57\enspace(1.01$\times$) \\

& WorldTrace-LM
& 47.56
& 73.05
& 61.12
& 53.15
& 57.99
& 76.58
& 62.69
& 62.57 & 7.67 & 29.85\enspace(0.76$\times$) \\

\cmidrule(lr){2-12}

& WorldKV
& 47.95
& 73.62
& 61.85
& 55.55
& 54.47
& 76.61
& 62.90
& 102.00 & 4.71 & 21.33\enspace(1.07$\times$) \\

& \textbf{DraftKV}
& \textbf{51.96}
& \textbf{74.06}
& \textbf{62.37}
& \textbf{58.98}
& \textbf{59.23}
& \textbf{78.65}
& \textbf{65.29}
& 87.51 & 5.49 & \textbf{18.56\enspace(1.23$\times$)} \\

\midrule

\multirow{7}{*}{\makecell{LingBot\\14B}}
& Baseline
& 60.19
& 74.49
& 61.44
& 57.72
& 57.21
& 78.67
& 67.04
& 191.58 & 2.51 & 62.36\enspace(1.00$\times$) \\

& Deep Forcing
& 60.32
& \underline{74.80}
& \underline{62.01}
& 60.00
& 57.22
& 78.91
& 67.49
& 213.70 & 2.25 & 126.24\enspace(0.49$\times$) \\

& Future Forcing
& 16.11
& 66.13
& 59.34
& 30.66
& 24.33
& 59.16
& 43.82
& 330.95 & 1.45 & 133.91\enspace(0.47$\times$) \\

& WorldTrace-LM
& 60.33
& 74.38
& 61.09
& 56.63
& 55.17
& 78.25
& 66.62
& 195.68 & 2.45 & 105.23\enspace(0.59$\times$) \\

\cmidrule(lr){2-12}

& WorldKV
& \underline{63.95}
& 74.54
& 61.90
& \underline{62.49}
& \underline{58.86}
& \underline{79.70}
& \underline{68.89}
& 344.13 & 1.40 & 99.36\enspace(0.63$\times$) \\

& \textbf{DraftKV}
& \textbf{66.34}
& \textbf{74.81}
& \textbf{62.55}
& \textbf{65.60}
& \textbf{60.38}
& \textbf{81.18}
& \textbf{70.38}
& 281.47 & 1.71 & \textbf{45.93\enspace(1.36$\times$)} \\

\midrule

\multirow{7}{*}{\makecell{SolarWM\\33B}}
& Baseline
& 84.90
& 74.15
& 68.97
& 72.86
& 67.09
& 86.91
& 78.04
& 360.41 & 1.33 & 118.22\enspace(1.00$\times$) \\

& Deep Forcing
& 84.66
& \underline{74.27}
& \underline{70.49}
& 71.85
& 66.46
& 86.91
& 78.00
& 387.16 & 1.24 & 188.05\enspace(0.63$\times$) \\

& Future Forcing
& 83.46
& 73.26
& 68.95
& 69.14
& 66.50
& 85.52
& 76.74
& 946.50 & 0.51 & 113.38\enspace(1.04$\times$) \\

& WorldTrace-LM
& \underline{85.36}
& 74.04
& 69.51
& \underline{74.68}
& \underline{67.46}
& \underline{87.09}
& \underline{78.43}
& 370.63 & 1.30 & 150.18\enspace(0.79$\times$) \\

\cmidrule(lr){2-12}

& WorldKV
& 84.86
& 74.25
& 70.00
& 73.11
& 67.11
& 87.02
& 78.20
& 1653.20 & 0.29 & 124.60\enspace(0.95$\times$) \\

& \textbf{DraftKV}
& \textbf{87.23}
& \textbf{74.54}
& \textbf{71.87}
& \textbf{78.78}
& \textbf{69.10}
& \textbf{88.41}
& \textbf{80.10}
& 504.84 & 0.95 & \textbf{72.05\enspace(1.64$\times$)} \\

\bottomrule
\end{tabular}%
}
\vspace{-4mm}
\end{table*}

\paragraph{Tile-Wise Readiness.}
Attention does not require this dependency. We process $H$ attention heads in groups of $G$ heads. 
For each head group with query $Q$, we divide the $N$ KV tokens into $M$ tiles $U_1,\ldots,U_M$, each containing at most $C$ tokens.
Tile $U_j$, with keys $K_{U_j}$ and values $V_{U_j}$, yields a partial output and its row-wise log-sum-exp (LSE),
\begin{equation}
\begin{gathered}
O_j=\operatorname{softmax}\!\Big(\tfrac{QK_{U_j}^{\top}}{\sqrt{d}}\Big)V_{U_j},\qquad
\lambda_j=\operatorname{LSE}\!\Big(\tfrac{QK_{U_j}^{\top}}{\sqrt{d}}\Big),\\
O=\sum_{j=1}^{M}e^{\lambda_j-\lambda}\,O_j,\qquad
\lambda=\log\sum_{j=1}^{M}e^{\lambda_j},
\end{gathered}
\label{eq:tile-merge}
\end{equation}
where $d$ is the head dimension and the exponentials act row-wise. Eq.~\eqref{eq:tile-merge} recovers full-window attention exactly, and the contribution of each tile depends only on its own KV, so attention on $U_j$ can start as soon as its KV has been transferred and prepared on the GPU. FlashAttention applies the same online-softmax reduction to KV blocks inside a kernel~\cite{flashattention}, and Ring Attention uses it to overlap KV-block exchange across devices with blockwise computation~\cite{ringattention}. SCG-Attention brings this property to query-dependent retrieval from CPU memory: it replaces selected-set readiness with tile-wise transfer, preparation, and attention through a bounded KV Reuse Buffer shared across head groups. 
FlashAttention independently computes the partial attention output and row-wise LSE for each tile, which are then combined through a final FP32-weighted merge using Eq.~\eqref{eq:tile-merge} to recover the full-window attention output.

\begin{table*}[!t]
\centering
\caption{
\textbf{Ablation of Hierarchical Draft Retrieval.}
We ablate Draft Grounding and Draft Refinement with different model families on ORBIA following the evaluation protocol of Table~\ref{tab:main_results}. The additional similarity results are included for further demonstration.
}
\label{tab:retrieval_ablation}
\vspace{-3mm}
\setlength{\tabcolsep}{4.9pt}
\renewcommand{\arraystretch}{1}
\small

\resizebox{\textwidth}{!}{%
\begin{tabular}{llcccccc|c|ccc}
\toprule

& &
\multicolumn{7}{c}{\textbf{Evaluation Scores}} &
\multicolumn{3}{c}{\textbf{Revisit Similarity}} \\

\cmidrule(lr){3-9}
\cmidrule(lr){10-12}

\textbf{Model} &
\textbf{Variant} &
\textbf{Ctrl.} $\uparrow$ &
\textbf{Qual.} $\uparrow$ &
\textbf{InPres.} $\uparrow$ &
\textbf{Persist.} $\uparrow$ &
\textbf{3D Cons.} $\uparrow$ &
\textbf{Stab.} $\uparrow$ &
\textbf{Overall} $\uparrow$ &
\textbf{PSNR} $\uparrow$ &
\textbf{LPIPS} $\downarrow$ &
\textbf{SSIM} $\uparrow$ \\

\midrule

\multirow{3}{*}{\makecell{LingBot\\1.3B}}
& w/o Draft Grounding
& 50.60 & 73.71 & 62.09
& 56.98 & 58.52 & 78.05 & 64.45
& 16.33 & 0.852 & 0.198 \\

& w/o Draft Refinement
& 50.98 & 74.09 & 62.19
& 57.75 & 58.61 & 78.63 & 64.85
& 16.27 & 0.845 & 0.204 \\

& \textbf{DraftKV}
& \textbf{51.95}
& \textbf{74.06}
& \textbf{62.37}
& \textbf{58.98}
& \textbf{59.23}
& \textbf{78.65}
& \textbf{65.29}
& 16.31 & \textbf{0.845} & 0.202 \\

\midrule

\multirow{3}{*}{\makecell{LingBot\\14B}}
& w/o Draft Grounding
& 60.81 & 74.78 & 62.13
& 60.33 & 57.35 & 78.96 & 67.67
& 15.24 & 0.834 & 0.219 \\

& w/o Draft Refinement
& 64.15 & 74.80 & 62.08
& 62.26 & 58.91 & 80.00 & 69.06
& 15.43 & 0.807 & 0.232 \\

& \textbf{DraftKV}
& \textbf{66.34}
& \textbf{74.81}
& \textbf{62.55}
& \textbf{65.60}
& \textbf{60.38}
& \textbf{81.18}
& \textbf{70.38}
& \textbf{15.85}
& \textbf{0.773}
& \textbf{0.250} \\

\midrule

\multirow{3}{*}{\makecell{SolarWM\\33B}}
& w/o Draft Grounding
& 85.65 & 74.25 & 70.63
& 75.43 & 68.04 & 87.62 & 78.91
& 17.49 & 0.708 & 0.303 \\

& w/o Draft Refinement
& 85.35 & 74.23 & 69.76
& 75.07 & 67.42 & 87.15 & 78.55
& 17.28 & 0.721 & 0.294 \\

& \textbf{DraftKV}
& \textbf{87.23}
& \textbf{74.54}
& \textbf{71.87}
& \textbf{78.78}
& \textbf{69.10}
& \textbf{88.41}
& \textbf{80.10}
& \textbf{17.95}
& \textbf{0.664}
& \textbf{0.329} \\

\bottomrule
\end{tabular}%
}
\vspace{-4mm}
\end{table*}

\begin{table}[!t]
\centering
\caption{
Comparison of revisit consistency with similarity metrics for videos generated by the methods in Table~\ref{tab:main_results}.
}
\label{tab:reprojection_result}
\vspace{-3mm}
\setlength{\tabcolsep}{4pt}
\renewcommand{\arraystretch}{1.12}
\small

\resizebox{0.9\columnwidth}{!}{%
\begin{tabular}{llccc}
\toprule
\textbf{Model} &
\textbf{Method} &
\textbf{PSNR $\uparrow$} &
\textbf{LPIPS $\downarrow$} &
\textbf{SSIM $\uparrow$} \\
\midrule

\multirow{6}{*}{\makecell{LingBot\\1.3B}}
& Baseline
& 16.21 & \underline{0.853} & \underline{0.198} \\

& Deep Forcing
& 16.23 & 0.864 & 0.184 \\

& Future Forcing
& \underline{16.27} & 0.861 & 0.187 \\

& WorldTrace-LM
& 15.41 & 0.899 & 0.176 \\

& WorldKV
& 15.71 & 0.888 & 0.174 \\
& \textbf{DraftKV}
& \textbf{16.31} & \textbf{0.845} & \textbf{0.202} \\

\midrule

\multirow{6}{*}{\makecell{LingBot\\14B}}
& Baseline
& 15.15 & 0.831 & 0.216 \\

& Deep Forcing
& 15.15 & 0.834 & 0.216 \\

& Future Forcing
& 12.50 & 1.118 & 0.078 \\

& WorldTrace-LM
& 14.70 & 0.870 & 0.197 \\

& WorldKV
& \underline{15.51} & \underline{0.793}
& \underline{0.238} \\
& \textbf{DraftKV}
& \textbf{15.85} & \textbf{0.773}
& \textbf{0.250} \\

\midrule

\multirow{6}{*}{\makecell{SolarWM\\33B}}
& Baseline
& 17.00 & 0.741 & 0.281 \\

& Deep Forcing
& 17.01 & 0.749 & 0.277 \\

& Future Forcing
& 17.09 & 0.786 & 0.246 \\

& WorldTrace-LM
& \underline{17.36} & \underline{0.717}
& \underline{0.296} \\

& WorldKV
& 17.07 & 0.733 & 0.285 \\
& \textbf{DraftKV}
& \textbf{17.95} & \textbf{0.664}
& \textbf{0.329} \\

\bottomrule
\end{tabular}%
}
\vspace{-3mm}
\end{table}

\paragraph{Streaming Pipeline.}
Algorithm~\ref{alg:streaming-cg} turns tile-wise readiness into a three-stage pipeline (Fig.~\ref{fig:framework}(c)). Each tile combines the historical KV fetched from CPU memory with the sink and current tokens it covers, is prepared at its assigned temporal positions, and is then consumed by attention. 
The three stages run asynchronously on separate CUDA streams and synchronize through per-tile completion flags in device memory, so while attention consumes one tile, subsequent tiles are already being transferred and prepared.
The exposed historical-access cost thus shrinks from the transfer of the entire retrieved set to pipeline fill and drain and the residual imbalance among the stages. 
Furthermore, we present more details of the memory layout, transfer planning, and synchronization in Appendix~\ref{sec:appendix-streaming}.

\paragraph{Bounded KV Reuse Buffer.}
Tile-wise readiness also bounds how much retrieved KV must reside on the GPU at once. Tiles land in a KV Reuse Buffer $\mathcal{B}$ of $R$ slots $\mathcal{B}_0,\ldots,\mathcal{B}_{R-1}$, each holding $C$ tokens of one head group, and the tiles of all head groups cycle through these slots in transfer order. A slot is refilled only after attention has finished reading its previous tile and stored the partial output and LSE, which keeps transfer at most $R$ tiles ahead of attention; with $R=2$, one slot is consumed while the other is filled. The buffer therefore occupies
\begin{equation}
|\mathcal{B}|=2\,R\,C\,G\,d\,b
\label{eq:bounded-buffer}
\end{equation}
bytes, where the factor $2$ counts K and V and $b$ is the number of bytes per element. A longer retrieved context only increases the number of tiles that stream through this buffer, whereas selected-set readiness materializes all $2NHd\,b$ bytes of the window. The active retrieved-KV residency on the GPU thus stays bounded with respect to the retrieved-context size, while full-resolution history remains in CPU memory and is searched through Draft K Bank.

\section{Experimental Results}

\begin{figure*}[t]
\vspace{-4mm}
    \centering
    \includegraphics[width=\textwidth]{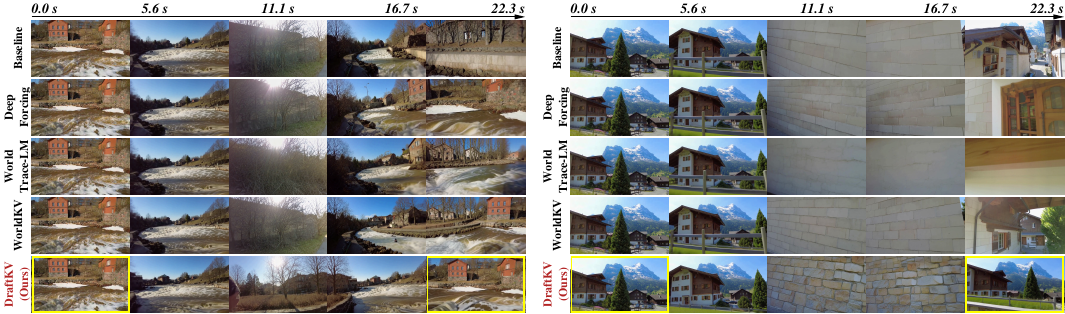}
    \vspace{-7mm}
    \caption{
    Comparison of long-horizon world persistence on ORBIA revisit cases. Complete prompts are provided in
    Appendix~\ref{app:qualitative_prompts}. 
    }
    \label{fig:revisit_qualitative}
    \vspace{-4mm}
\end{figure*}

\subsection{Experiment Setup}
We evaluate DraftKV on 60\,s VBench-Long~\cite{vbenchlong} with LongLive-1.3B~\cite{longlive}, and on all 480 ORBIA trajectories~\cite{orbia} with LingBot-World 1.3B/14B~\cite{lingbotworld} and SolarWM-33B~\cite{solarwm}. We compare against native inference and representative KV-management baselines, including WorldKV, FreeKV~\cite{freekv}, InfLLM~\cite{infllm}, InfiniGen, ShadowKV, WorldTrace-LM, Deep Forcing~\cite{deepforcing}, and Future Forcing~\cite{futureforcing}, using identical inputs, native attention geometry, and matched active-context budgets on ORBIA. We report VBench-Long quality metrics, ORBIA scores, revisit consistency with PSNR/SSIM~\cite{ssim}/LPIPS~\cite{lpips}, and efficiency in end-to-end latency, throughput, and peak GPU memory. More setup details see in Appendix.~\ref{sec:appendix-protocol}.

\subsection{Main Results}

\textbf{Generation quality.}
DraftKV preserves generation quality while introducing long-horizon historical retrieval.
On 60\,s VBench-Long, it achieves a quality score of 82.08, matching the native baseline (82.07) and remaining competitive across individual quality dimensions (Table~\ref{tab:vbench_results}).
The same trend holds on ORBIA, where quality, input preservation, and stability remain comparable to or improve over native inference across all three backbones (Table~\ref{tab:main_results}).
These results indicate that historical retrieval does not substantially disrupt the pretrained model's native generation capability.

\noindent\textbf{World persistence.}
DraftKV consistently improves persistence and 3D consistency across all three ORBIA backbones (Table~\ref{tab:main_results}), with the largest persistence gain on LingBot-14B from 57.72 to 65.60 and the best overall score on all three models.
Direct revisit evaluation in Table~\ref{tab:reprojection_result} further shows stronger agreement between revisited frames and their corresponding first-visit observations on most model--metric combinations.
Together, these results show that the aggregate persistence gains translate into better recovery of previously observed appearance and scene structure, as also visualized in Fig.~\ref{fig:revisit_qualitative}.

\noindent\textbf{Efficiency.}
DraftKV also provides a favorable quality--latency--memory trade-off.
On SolarWM-33B, CPU-offloaded DraftKV is $3.27\times$ faster than WorldKV while reducing GPU memory from 124.60 to 72.05\,GiB.
These gains show that historical KV retrieval remains practical at large model scales despite CPU offloading.

\begin{figure}[t]
    \centering
    \vspace{-3mm}
    \includegraphics[width=\linewidth]{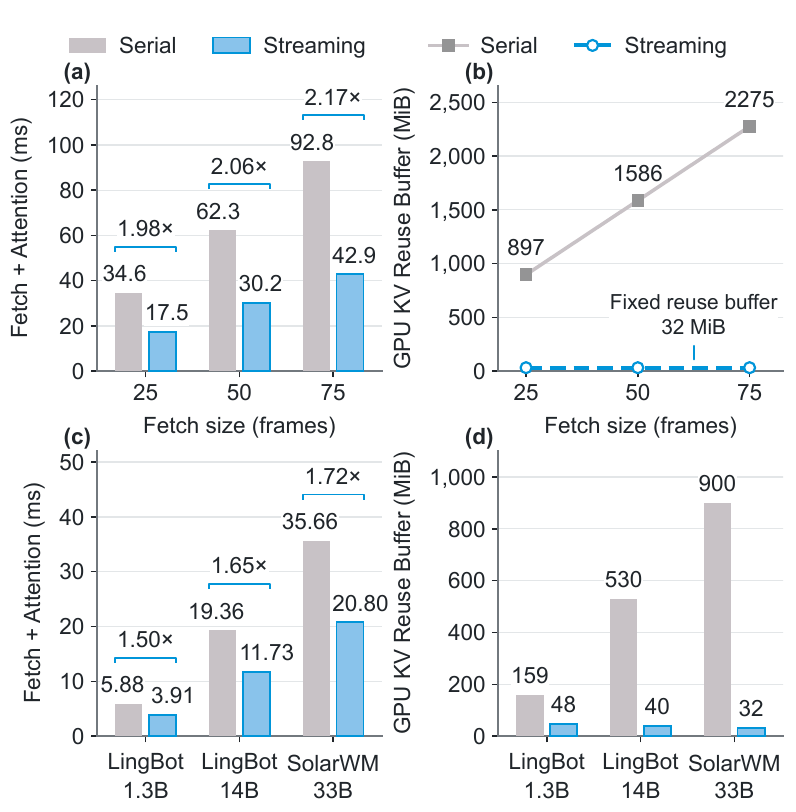}
    \vspace{-6mm}
    \caption{
    Scaling behavior of Streaming CPU-to-GPU Attention.
    (a--b) Fetch+Attn latency and GPU KV reuse-buffer memory as the retrieved context increases on SolarWM-33B.
    (c--d) Corresponding latency and memory measurements across different model scales.
    }
    \label{fig:streaming_scaling}
    \vspace{-4mm}
\end{figure}

\subsection{Ablation Study}

\noindent\textbf{Hierarchical Draft Retrieval.}
Table~\ref{tab:retrieval_ablation} shows that removing either Draft Grounding or Draft Refinement consistently degrades persistence and 3D consistency across all three backbones.
Meanwhile, we present the visualization results in Fig.~\ref{fig:retrieval_motivation}.
Overall, the full model attains the best score on all three backbones, supporting the complementary ground-then-refine design in Sec.~\ref{sec:theoretical-analysis}.

\noindent\textbf{SCG-Attention.}
We isolate SCG-Attention by fixing the retrieved KV and varying only its transfer and execution.
Figure~\ref{fig:streaming_scaling} shows consistent scaling with both retrieved context and model size.
Across model scales, SCG-Attention achieves a $1.50$--$1.72\times$ Fetch+Attn speedup while reducing the active KV reuse buffer from 159--900,MiB to 32--48,MiB.
As the retrieved context grows, serial execution incurs increasing transfer and selected-set materialization overhead, whereas Streaming overlaps H2D transfer with tile-wise attention while keeping the reuse buffer fixed.
These trends validate the tile-wise execution in Sec.~\ref{sec:streaming-cg}: increasing retrieved history adds streamed tiles rather than increasing the amount of retrieved KV simultaneously resident on the GPU.

\noindent\textbf{Retrieval granularity.}
Under the same active KV budget, block-level retrieval achieves 98.01\% Recall@$K$, compared with 88.87\% for the best frame-level variant, demonstrating the benefit of fine-grained refinement under spatially non-uniform relevance (Appendix~\ref{sec:appendix-granularity}).

\noindent\textbf{Retrieval-signal validation.}
Draft Grounding achieves 95.42\% observation Recall@4 over 480 ORBIA trajectories. Draft Block Relevance achieves 0.990 Spearman correlation and 95.62\% Recall@$K$ against Full-Resolution Block Relevance.
(Complete analyses in Appendix~\ref{sec:appendix-retrieval}.)

\section{Conclusion}
We introduced DraftKV, a training-free framework for efficient historical KV-cache retrieval in long-horizon video world models through hierarchical draft retrieval and streaming attention. 
Our theoretical analysis establishes draft relevance approximation and retrieval utility bounds, and justifies geometric grounding for spatial consistency.
Experiments show that, 
DraftKV improves generation quality and revisit consistency across model scales and benchmarks while reducing GPU memory and retrieval latency through CPU-offloaded KV streaming.

\section*{LLM Usage Statement}
OpenAI (ChatGPT / GPT), Google (Gemini) and DeepSeek were used to assist with writing and editing text, preparing related work and citations, code implementation, experimental design and analysis, data preparation, and refining research ideas and conclusions. No generative AI tools were used to create figures or other visual materials. All content was reviewed and verified by the authors.

{
    \small
    \bibliographystyle{ieeenat_fullname}
    \bibliography{reference}
}

\clearpage
\setcounter{page}{1}
\maketitlesupplementary

\appendix

\section{Theoretical Supplementary}
\label{sec:appendix-theory}

This appendix proves the remaining results of
Sec.~\ref{sec:theoretical-analysis}. Throughout we write interactions for one
attention head and fix one generated chunk $t$, omit the head index where it
plays no role, and write
$\mathcal{B}_t=\mathcal{B}(\mathcal{C}_t\cup\mathcal{P}_t)$ for the scoring
domain over which both relevance scores are normalized. Competitive retrieval
is restricted to
$\mathcal{B}^{\mathrm{ret}}_t=\mathcal{B}(\mathcal{C}_t)$.

\paragraph{Proof of Lemma~\ref{lem:pooling}.}
By bilinearity of the inner product,
\begin{equation*}
  \frac{1}{|\mathcal{Q}_r||\mathcal{K}_b|}\sum_{i\in\mathcal{Q}_r}\sum_{j\in\mathcal{K}_b}
  \frac{\langle q_i,k_j\rangle}{\sqrt{d}}
  =\frac{\langle\bar q_r,\bar k_b\rangle}{\sqrt{d}},
\end{equation*}
so $\mu^{(h)}_{r,b}=\widehat{\mu}^{(h)}_{r,b}$ for every head and every pair
$(r,b)$, with no approximation. The mean-branch logits therefore coincide, and
since both scores normalize them over the same domain $\mathcal{B}_t$, the
probabilities $P^{(h),\mathrm{mean}}_{r,b}$ coincide as well. Subtracting
Eq.~\eqref{eq:importance} from its draft counterpart leaves only the max
branch,
\begin{equation*}
  \widehat{I}_b-I_b
  =\frac{\beta}{HN_q}\sum_{h,r}
   \Big(\widehat{P}^{(h),\max}_{r,b}-P^{(h),\max}_{r,b}\Big),
\end{equation*}
and each probability lies in $[0,1]$, so $|\widehat{I}_b-I_b|\le\beta$.
\hfill$\square$

\paragraph{Ratio bound for pooled descriptors.}
For any head $h$ and region $r$, Cauchy--Schwarz gives
$|\widehat{\mu}^{(h)}_{r,b}-\widehat{\mu}^{(h)}_{r,b'}|
 =|\langle\bar q^{(h)}_r,\bar k^{(h)}_b-\bar k^{(h)}_{b'}\rangle|/\sqrt{d}\le\eta$,
and likewise
$|\widehat{X}^{(h)}_{r,b}-\widehat{X}^{(h)}_{r,b'}|\le\eta$. The two blocks
share the same softmax normalizer, so each of
$\widehat{P}^{(h),\mathrm{mean}}_{r,b}/\widehat{P}^{(h),\mathrm{mean}}_{r,b'}$
and $\widehat{P}^{(h),\max}_{r,b}/\widehat{P}^{(h),\max}_{r,b'}$ lies in
$[e^{-\eta},e^{\eta}]$. Since $\widehat{I}_b$ is a non-negative combination of
these terms, the same two-sided bound holds for
$\widehat{I}_b/\widehat{I}_{b'}$.
\hfill$\square$

\paragraph{Proof of Proposition 3.4.}
Let $p$ be a scene point observed by block $b$ of frame $s$ and by region $r$
of frame $t$. Being observed by frame $s$ means $p$ lies on a pixel ray of $s$
at a depth inside the visible range, that is, $p\in\mathcal{F}_s$; likewise
$p\in\mathcal{F}_t$. Hence $p\in\mathcal{F}_s\cap\mathcal{F}_t$ and the
intersection is non-empty. The scene is static, so a co-visible
correspondence between $b$ and $r$ is by definition the image of such a shared
point. If the retained set contains every frame $s$ with
$\mathcal{F}_s\cap\mathcal{F}_t\neq\emptyset$, every frame carrying such a
correspondence is kept, and therefore so is every block of it.
\hfill$\square$

\paragraph{From the idealized rule to Draft Grounding.}
Proposition~3.4 considers exact frame-level frustum
intersections. In practice, Draft Grounding uses chunk-level camera estimates
and sampled bidirectional overlap scores, followed by budgeted candidate
selection. Sink chunks and recent historical chunks are excluded from
competitive retrieval; protected history is handled separately, leaving an
eligibility gap near its boundary. Thus, the ideal co-visibility guarantee
does not directly extend to the implemented retrieval policy.

\section{Streaming CG-Attention Implementation Details}
\label{sec:appendix-streaming}

\paragraph{Memory layout and transfer.}
After the clean forward pass of each generated chunk, the mean- and max-pooled Draft K descriptors of its spatial blocks are appended to the GPU-resident Draft K Bank, and its full-resolution KV is committed to pinned host memory. The archive stores each spatial block as consecutive head groups with contiguous K and V payloads, so a selected block maps to one K span and one V span per head group. Keys keep their raw temporal components, so temporal K-RoPE can be applied directly at the assigned positions. After retrieval, the GPU converts the selected block IDs into DMA descriptors, and the host only submits the spans of each tile as one batched copy, without gathering or repacking KV on the CPU.

\paragraph{Tiling and synchronization.}
A head-group window of $N$ tokens is split into $J=\min\!\left(N,\max(S,\lceil N/C\rceil)\right)$ contiguous tiles, where $S$ is the requested number of partitions per head group and $C$ is the tile capacity; tiles are issued head group by head group and share the $R$ slots in this order. Within a head group, the configured schedule can issue tiles in reverse order. Transfer, preparation, and attention run on three CUDA streams and are linked by per-tile completion flags in device memory, which the streams publish and await without host round trips. A slot is refilled only after the consumed flag of its previous tile is set, \ie, after attention has stored that tile's partial output and LSE.

The bounded executor tiles a disjoint union of the optional fixed prefix,
sink tokens, selected history (including protected history), and current
tokens. Historical DMA fills only the historical interval; views or copies
supply the complementary ranges. Borrowing a complete resident CUDA range
replaces its slot-backed read rather than adding another partition.
In the older head-group executor, the SolarWM condition/audio prefix is
instead attended once as a separate partition in the same LSE-weighted merge.

\paragraph{Reuse-buffer sizing.}
The GPU reuse buffer stores only the KV tiles currently staged for attention and is therefore determined by the tile capacity and head-group width, rather than by the total number of attention heads or the retrieved-context length.
Here $S$, $R$, and $C$ are execution parameters, distinct from the main-text context-budget notation.
For $R$ reusable slots, tile capacity $C$, head-group width $G$, head dimension $d$, and element size $b$, the buffer size is
\begin{equation}
    M_{\mathrm{buf}}
    =
    R \cdot 2 \cdot C \cdot G \cdot d \cdot b ,
    \label{eq:appendix-buffer-sizing}
\end{equation}
where the factor $2$ accounts for K and V.

\begin{table}[!t]
\centering
\caption{
\textbf{SP1 receive-buffer configuration across model scales.}
Larger models contain more attention heads, but head groups are processed sequentially and reuse the same slots.
Thus, the receive-buffer capacity is determined by the configured head-group width $G$, rather than the total number of heads.
}
\label{tab:appendix-buffer-sizing}

\setlength{\tabcolsep}{2pt}
\renewcommand{\arraystretch}{1.10}
\normalsize

\begin{tabular}{lrrrr}
\toprule
\textbf{Model} &
\textbf{Heads} &
\textbf{$G$} &
\textbf{Groups} &
\textbf{Buffer (MiB)} \\
\midrule
LingBot-1.3B & 12 & 12 & 1 & 48 \\
LingBot-14B  & 40 & 10 & 4 & 40 \\
SolarWM-33B  & 56 & 8  & 7 & 32 \\
\bottomrule
\end{tabular}
\end{table}

Importantly, ``fixed'' means fixed with respect to the retrieved-context length, rather than identical across models.
Although larger models contain more attention heads, different head groups are streamed sequentially and reuse the same slots; consequently, a smaller $G$ can yield a smaller reuse buffer even for a larger model.

\paragraph{Merge and configuration.}
After attention has stored all tile partial outputs and LSE values, a separate Triton kernel merges these stored partials with FP32 weights according to Eq.~\eqref{eq:tile-merge}.
For a single-batch query of shape $(1,Q,H,d)$ and $J>1$ external
partitions, the stored partial outputs and FP32 LSE have shapes
$(J,1,Q,H,d)$ and $(J,1,H,Q)$, requiring $JQH(db+4)$ bytes.
This excludes the final output, receive slots, and native-attention
temporaries; at $J=1$, the partial output aliases the final output.
Model weights, resident KV, and the persistent Draft K Bank are separate allocations.
Since $J$ can grow with the attended window, the bounded-memory guarantee
applies specifically to the active KV Reuse Buffer.
The execution configuration is selected for the target GPU architecture.
In the controlled selected-KV comparisons of Appendix~\ref{sec:appendix-streaming-ablation}, the Serial reference uses the same selected KV, positions, and attention partitions as Streaming, but launches attention only after all transfers and preparation have completed.

\paragraph{Independent attention correctness.}
We additionally compare Serial and Streaming with an independent FP32 dense
$QK^{\top}$--softmax--$V$ reference under matched token ordering and temporal
positions. Across 244 controlled numerical checks, including production head-group
sizes, the largest BF16 output relative $L_2$ error is $2.84\times10^{-3}$.
All 48 equal-partition Serial/Streaming pairs are bitwise equal. These are
numerical/positional checks rather than complete-generation experiments.

\begin{table}[t]
\centering
\normalsize
\caption{\textbf{Independent dense-attention correctness checks.}
The independent FP32 dense reference is compared with BF16 Serial/Streaming execution on a controlled positional/tiling matrix, including production head-group sizes. Maxima are measured across the tested cases; equal-partition Serial/Streaming pairs agree bitwise, whereas FP32/BF16 outputs are only numerically close.}
\label{tab:app_dense_correctness_new}
\setlength{\tabcolsep}{4pt}
\begin{tabular}{lr}
\toprule
Check & Verified result \\
\midrule
Passed correctness checks & 244 / 244 \\
Max output relative $L_2$ error & $2.84\times 10^{-3}$ \\
Max absolute output error & $5.52\times10^{-4}$ \\
Max LSE error & $1.91\times10^{-6}$ \\
Bitwise-equal Serial/Streaming pairs & 48 / 48 \\
\bottomrule
\end{tabular}
\end{table}

\section{Experimental Details and Qualitative Material}
\label{sec:appendix-exp-details}

\subsection{Evaluation Protocols and Pose-Free Configuration}
\label{sec:appendix-protocol}
VBench-Long follows the native LongLive-1.3B protocol for 60\,s videos at
16\,FPS. ORBIA's short-evaluation protocol contains 480 trajectories,
each targeting 357 decoded frames at 16\,FPS for LingBot-World 1.3B/14B
and SolarWM-33B. The count of 480 here denotes trajectories, not frames.
Main-paper Table~2 reports ORBIA evaluation scores, whereas its E2E
latency, throughput (FPS), and peak GPU memory are measured on separate
30\,s, 480-frame WorldAtlas system cases. These are distinct workloads:
Table~2's E2E/FPS must not be interpreted as measurements of the
357-frame ORBIA evaluation videos. Corpus quality scores aggregate finite leaf
scores first, using each leaf's own valid denominator, before applying the
original hierarchy. Overall is computed from the leaves available in the
ORBIA short-evaluation protocol, consistent with main-paper Table~2;
metrics exclusive to the complete protocol are not included. Revisit evaluation
follows the source-locked reconstruction and reprojection path at 3\,FPS
with its original visibility mask. PSNR retains native valid-pixel and
positive-score aggregation; SSIM uses masked Gaussian windows, and spatial
LPIPS uses AlexNet features on patches. Windows and patches require the
original valid-pixel support. These masked correspondence metrics and
quality scores retain their respective finite sample supports.

For pose-free text-only generation, camera grounding is unavailable and
camera-grounding options are rejected. Retrieval considers clean-committed
non-sink history, protects recent historical context, and directly ranks
the remaining blocks. Camera-specific temporal eligibility is not imposed
on this route. Selected tokens retain the native positional policy and
are consumed by Streaming CG-Attention; camera-controlled correspondence
results are reported separately.

Complete-generation Serial/Streaming controls use matched inputs, seeds,
selected KV, shared resident routes, positions, temporal K-RoPE, and
attention partitions. Timing starts at the synchronized generation call
after model loading and extension preparation. LingBot includes conditioning
and VAE decoding but excludes video encoding. SolarWM reuses verified
conditioning and includes native decoding and video writing. Fresh
processes use the stated warm compilation-cache policy; first-use runtime
initialization remains inside the timer and is distinct from a compiler
invocation. NVML peaks are sampled device/process measurements, distinct
from PyTorch peak allocated tensor memory. Fixed-query measurements and deployment-capability checks
retain their separately stated scopes. Raw seconds across backbones do
not represent a common workload.

Across retrieval methods, comparisons retain their native retention and
compression policies within the audited cohorts. Realized token counts
can differ with compression or dynamic selection, so these configured-system
comparisons do not isolate a single factor. The controlled selected-KV
Serial/Streaming comparisons provide the execution-specific evidence.

\subsection{Qualitative Example Prompts}
\label{sec:appendix-prompts}
\label{app:qualitative_prompts}

We provide the complete text prompts for the two
SolarWM-33B qualitative examples shown in
Fig.~\ref{fig:revisit_qualitative}.
The left example corresponds to case\_0006 (Riverside),
while the right example corresponds to case\_0374
(Mountain Village).

\paragraph{Riverside (Left, case\_0006).}
A scene where a turbulent, brown river flows rapidly
past a stone embankment, carrying patches of melting
snow. The water churns with white foam as it moves
downstream. On the bank, a prominent red brick
building with white-framed windows and a stone
foundation stands adjacent to another similar structure.
The terrain is a mix of grassy slopes and rocky ground,
dotted with bare trees and shrubs. The lighting is bright
and sunny, casting distinct shadows and highlighting
the textures of the brickwork and the rushing water.

\paragraph{Mountain Village (Right, case\_0374).}
A scenic mountain village scene where traditional
wooden chalets with white plastered lower levels
and brown timber upper stories stand alongside a
large, dark coniferous tree. The architecture features
numerous windows with wooden shutters and steep,
tiled roofs. In the distance, a massive, rugged mountain
peak rises, its upper slopes covered in snow and
glaciers, while the lower slopes are densely forested
with green trees. The terrain is hilly and grassy,
dotted with other buildings and a paved road with
parked cars. The atmosphere is bright and clear
under a blue sky, with sunlight illuminating the
landscape and casting soft shadows.

\subsection{Additional Ablation Visualizations}
\label{sec:appendix-ablation-vis}

\begin{figure*}[!tp]
    \centering
    \includegraphics[width=\textwidth,keepaspectratio]{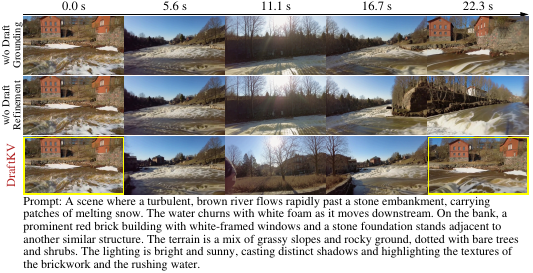}
    \caption{
    \textbf{Visualization of ablation algorithms on ORBIA revisit cases.}
    }
    \label{fig:revisit_qualitative_ablation}
\end{figure*}

Figure~\ref{fig:revisit_qualitative_ablation} presents additional qualitative ablations on ORBIA revisit cases.

\begin{figure*}[!tp]
    \centering
    \includegraphics[width=\textwidth,keepaspectratio]{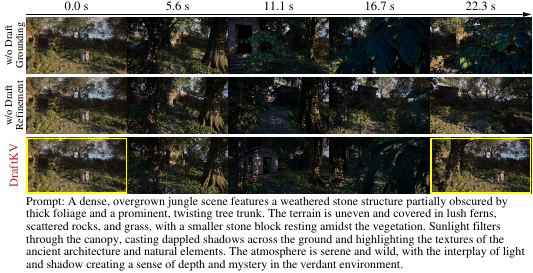}
    {\par\vspace{4pt}\small\noindent\textbf{Figure~\ref{fig:revisit_qualitative_ablation} (continued).}\par}
\end{figure*}

\subsection{Additional Qualitative Comparisons}
\label{sec:appendix-qualitative-comparisons}

\begin{figure*}[!tp]
    \centering
    \includegraphics[width=0.94\textwidth,height=0.82\textheight,keepaspectratio]{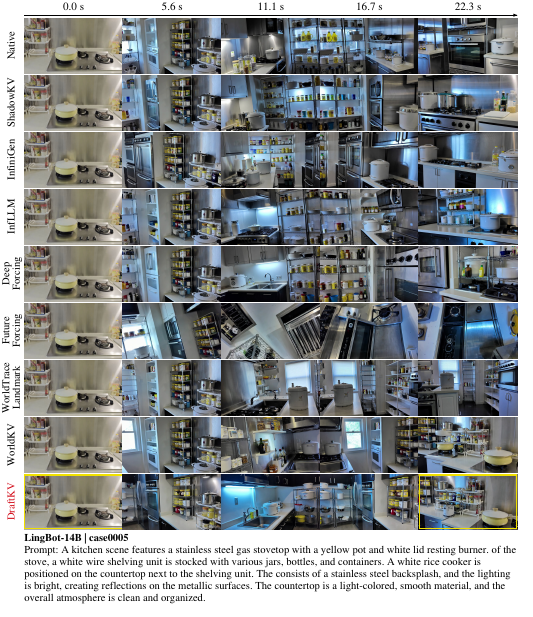}
    \caption{
    \textbf{Nine-method qualitative comparisons with LingBot-14B.}
    Kitchen (left, case\_0005) and woodland (right, case\_0003), comparing
    Native, ShadowKV, InfiniGen, InfLLM, Deep Forcing, Future Forcing,
    WorldTrace-Landmark, WorldKV, and DraftKV. The kitchen highlights
    the appearance and multiplicity of the yellow pot; the woodland
    highlights the stone landmark and surrounding scene composition.
    }
    \label{fig:appendix-qualitative-lingbot14b}
\end{figure*}

Figures~\ref{fig:appendix-qualitative-lingbot14b}--\ref{fig:appendix-qualitative-playground}
show five additional ORBIA cases using LingBot-14B and SolarWM-33B.
Each row samples the same five frame indices within a case: 0, 89, 178,
267, and 356 at 16\,FPS. The displayed timestamps are rounded to one
decimal place. Frames retain their complete field of view, and the original
text prompt appears below each panel. Yellow borders identify the first
and final DraftKV frames for visual comparison. The examples retain each
method's own retention and compression policy, following the comparison
scope in Sec.~\ref{sec:appendix-protocol}.

\paragraph{Object appearance and return-view composition.}
The kitchen example in Fig.~\ref{fig:appendix-qualitative-lingbot14b}
compares nine methods. DraftKV returns to a single yellow lidded pot and
the adjacent spice rack, whereas WorldKV shows two yellow pots in the
final sampled view. Future Forcing also retains a yellow pot, with a
substantially rotated view. The woodland example complements this object-level
comparison: DraftKV recovers a view resembling the initial stone landmark
and surrounding trees, while the alternatives exhibit different return-view
compositions. These selected cases illustrate appearance and layout
consistency alongside the aggregate evaluation.

\begin{figure*}[!tp]
    \centering
    \includegraphics[width=0.94\textwidth,height=0.82\textheight,keepaspectratio]{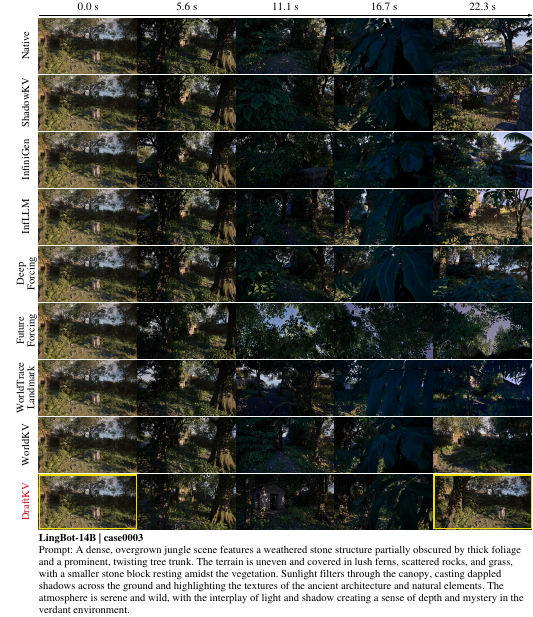}
    {\par\vspace{4pt}\small\noindent\textbf{Figure~\ref{fig:appendix-qualitative-lingbot14b} (continued).}\par}
\end{figure*}

\paragraph{Indoor and outdoor layouts.}
Figure~\ref{fig:appendix-qualitative-solarwm33b} compares five methods on
an indoor room and an outdoor street. In the room, DraftKV recovers the
orange lips artwork and nearby white door in the final sampled view.
In the street, the archway, wall poster, and bench provide recognizable
layout anchors. 
\begin{figure*}[!tp]
    \centering
    \includegraphics[width=\textwidth,keepaspectratio]{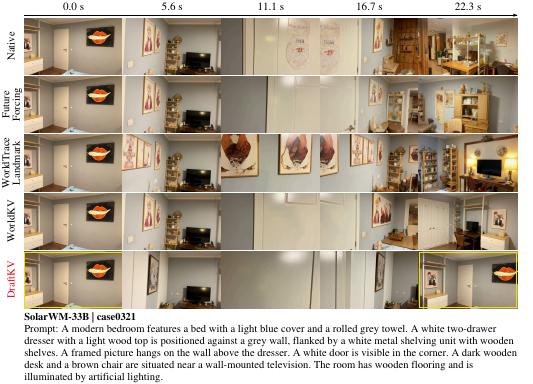}
    \caption{
    \textbf{Five-method qualitative comparisons with SolarWM-33B.}
    Room (left, case\_0321) and street (right, case\_0012), comparing
    Native, Future Forcing, WorldTrace-Landmark, WorldKV,
    and DraftKV. The examples show differences in the recovered artwork,
    doorway, street landmarks, and viewing direction after camera motion.
    }
    \label{fig:appendix-qualitative-solarwm33b}
\end{figure*}
\begin{figure*}[!tp]
    \centering
    \includegraphics[width=\textwidth,keepaspectratio]{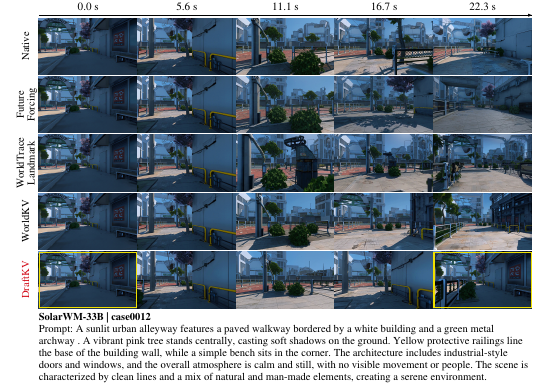}
    {\par\vspace{4pt}\small\noindent\textbf{Figure~\ref{fig:appendix-qualitative-solarwm33b} (continued).}\par}
\end{figure*}

\begin{figure*}[!tp]
    \centering
    \includegraphics[width=\textwidth,keepaspectratio]{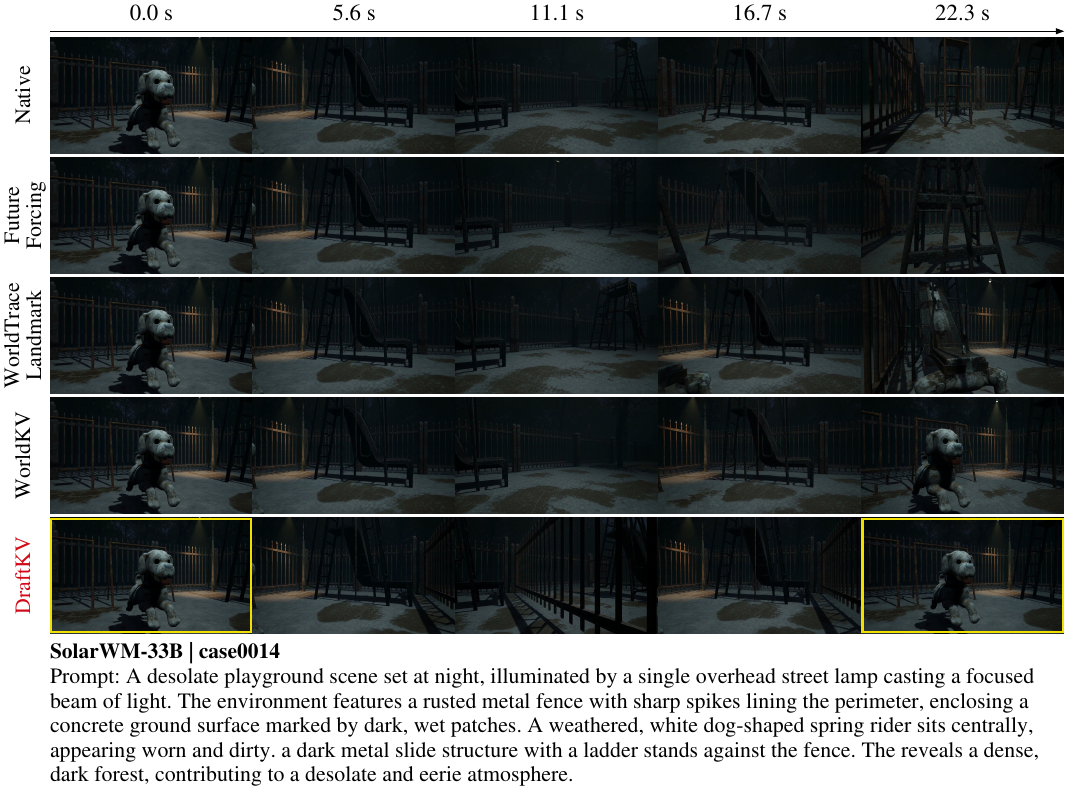}
    \caption{
    \textbf{Playground comparison with SolarWM-33B (case\_0014).}
    DraftKV and WorldKV both show the playground toy in
    their final sampled views. This example complements the room and
    street cases with a small foreground object.
    }
    \label{fig:appendix-qualitative-playground}
\end{figure*}

\section{Retrieval Analysis}
\label{sec:appendix-retrieval}
\label{sec:appendix-retrieval-analysis}
\label{sec:appendix-experiments}

We examine the roles of geometry-ranked admissibility and query-native relevance in Hierarchical Draft Retrieval. The original retrieval diagnostics reuse archived scores and metadata; additional validation experiments state their own evaluation scope and provenance. Unless otherwise noted, summaries weight trajectories equally, and repeated layer--step--query states are descriptive measurements within those trajectories. Protected $\mathcal P_t$ participates in the original normalization but is excluded from competitive retrieval statistics.

\subsection{Draft Grounding Analysis}
\label{sec:appendix-grounding}
\label{sec:appendix-grounding-analysis}

Draft Grounding narrows the history before content-based refinement.
In an observation-level LingBot candidate-ranking diagnostic, the designated
revisit observation is retained within the top three candidates for $442/480$
queries (Recall@$3$: $92.08\%$); the top-four cutoff yields $95.42\%$. Table~\ref{tab:app_grounding_robustness_new}
provides the additional candidate-count diagnostics; MRR is 0.740.
These observation-level labels are independent of the appearance diagnostic
below and do not supply a SolarWM recall estimate.

\begin{table}[t]
\centering
\caption{Grounding coverage and within-candidate localization on three matched SolarWM-33B cases (24 queries, 72 repeated layer states). PSNR is a descriptive frame-level revisit-similarity target, not ground-truth world correspondence. Rates are case-balanced descriptive means.}
\label{tab:app_grounding_coverage}
\normalsize
\setlength{\tabcolsep}{2pt}
\renewcommand{\arraystretch}{1.10}
\begin{tabular}{@{}p{0.56\columnwidth}r@{}}
\toprule
Diagnostic & Value \\
\midrule
Matched cases / queries & 3 / 24 \\
Eligible PSNR peak retained & 22/24 (91.67\%) \\
Top-3 / Top-5 PSNR coverage & 90.28\% / 92.50\% \\
Peak BFO rank: median / P90 & 2 / 6.7 \\
Historical search-space reduction & 31.44\% \\
Within-$C$ peak rank: mean / median & 13.04 / 11 \\
Within-$C$ exact / $\pm1$-frame match & 9.72\% / 18.06\% \\
Peak retained: 0005 / 0013 / 0051 & 7/8 / 7/8 / 8/8 \\
\bottomrule
\end{tabular}
\end{table}

For three matched SolarWM-33B cases, Table~\ref{tab:app_grounding_coverage} uses the actual saved candidate sets and defines the PSNR target only over production-eligible competitive history. The two missed peaks are case0005/q10/F4 and case0013/q14/F17. Cases 0172, 0689 and 0719 lack matched authoritative SolarWM PSNR/log pairs and are not imputed. Candidate admission (22/24 eligible PSNR peaks retained), ranking within the admitted set, and exact or near-exact PSNR-peak localization are distinct diagnostics. PSNR-max retention concerns descriptive appearance, not physical world correspondence; within-$C$ peak alignment is substantially less frequent than candidate admission.

\begin{table}[t]
\centering
\caption{Zero-score allocation at the production-equivalent block cap. $S_{\rm BFO}=0$ denotes zero score under the sampled BFO estimator, not exact scene-point nonintersection or proof of physical non-covisibility. Relevance-only ranks all captured non-sink, non-protected history; eligible-only additionally enforces production temporal eligibility. Fractions are trajectory means.}
\label{tab:app_sbfo0}
\normalsize
\setlength{\tabcolsep}{3pt}
\renewcommand{\arraystretch}{1.10}
\begin{tabular}{lrr}
\toprule
Ranking / quantity & 14B & 33B \\
\midrule
Relevance only: blocks & 5.74\% & 6.24\% \\
Relevance only: frames & 5.13\% & 7.74\% \\
Relevance only: mass & 5.94\% & 5.44\% \\
States with zero-score selection & 30/150 & 20/72 \\
Trajectories with zeros & 1/5 & 1/3 \\
Eligible-only: blocks & 6.09\% & 8.72\% \\
Eligible-only: mass & 6.21\% & 8.07\% \\
Actual hierarchy: blocks & 0.00\% & 5.27\% \\
Actual hierarchy: mass & 0.00\% & 4.87\% \\
Hierarchy states with zeros & 0/150 & 9/72 \\
\bottomrule
\end{tabular}
\end{table}

Table~\ref{tab:app_sbfo0} isolates the camera constraint by inspecting relevance-only selection at the same block cap. Zero-score events occur in one of five LingBot-14B trajectories and one of three SolarWM-33B trajectories. They remain after restricting the ranking to eligible history. The sampled bilateral score is a harmonic mean, so zero requires at least one directional sampled overlap to be zero, not both. Moreover, Top-$M$ is not a positivity threshold: the sampled SolarWM hierarchy includes zero-score blocks in nine states (case0005, queries 6, 8 and 10) when positive-score chunks are insufficient. These observations support Grounding as a coarse geometry-ranked admissibility stage, not as a complete geometric guarantee.

\paragraph{Candidate ranking and pose perturbations.}
Among the 38 queries missed at the top-three cutoff, 37 have at least
three candidates with strictly higher BFO scores and one is a cutoff tie.
Top-$M$ is a budgeted ranking, not a guarantee of preserving all truly
co-visible history. Table~\ref{tab:grounding_pose_noise_m3} reports
Recall@$3$ under camera-pose perturbations on the same 480 revisit queries.
Recall remains close to the unperturbed value under mild perturbations,
while larger pose errors progressively reduce candidate recall.
These are camera-ranking measurements rather than generated-video quality
measurements. For the separate Recall@$4$ diagnostic, the 22 misses comprise
12 arc, 7 yaw, 2 lateral, and 1 orbit cases.

\begin{table}[t]
\centering
\normalsize
\caption{\textbf{Candidate-count diagnostic for Draft Grounding.}
All candidate-count cutoffs use the same 480 ORBIA revisit queries.}
\label{tab:app_grounding_robustness_new}
\setlength{\tabcolsep}{8pt}
\begin{tabular}{cc}
\toprule
$M$ & Recall@$M$ (\%) $\uparrow$ \\
\midrule
1 & 54.17 \\
2 & 87.92 \\
3 & 92.08 \\
4 & 95.42 \\
6 & 97.71 \\
8 & 99.38 \\
12 & 99.79 \\
\bottomrule
\end{tabular}
\end{table}

\begin{table}[t]
\centering
\normalsize
\caption{\textbf{Pose-noise robustness of the top-three candidate-ranking diagnostic.}
Translation magnitude is relative to a trajectory-derived reference scale,
rather than in metric units. Each perturbation level uses the same 480
revisit queries; results average three perturbation seeds, which are repeated
measurements of those queries rather than additional independent samples.}
\label{tab:grounding_pose_noise_m3}
\setlength{\tabcolsep}{5pt}
\begin{tabular}{lcc}
\toprule
Perturbation & Magnitude & Recall@$3$ (\%) $\uparrow$ \\
\midrule
Rotation & $0^\circ$ & 92.08 \\
 & $2^\circ$ & 92.15 \\
 & $5^\circ$ & 91.60 \\
 & $10^\circ$ & 88.26 \\
 & $20^\circ$ & 74.86 \\
\midrule
Translation & $0\times$ & 92.08 \\
 & $0.25\times$ & 92.50 \\
 & $0.5\times$ & 91.53 \\
 & $1\times$ & 87.99 \\
 & $2\times$ & 72.57 \\
\bottomrule
\end{tabular}
\end{table}

\paragraph{Geometry--relevance selection diagnostic.}
A separate SolarWM-33B replay compares selected historical blocks under
matched token budgets on three trajectories (72 repeated states).
Geometry-first selection reduces sampled-zero-overlap allocation relative to
Draft-only retrieval, while discarding Full-Resolution-relevance proxy blocks
outside its candidate domain (Table~\ref{tab:app_geometry_draft_diagnostic_new}).
The proxy measures relevance rather than independent physical correspondence
or attention-output utility. The reverse-order control also changes geometric
eligibility and normalization, so it is not a single-factor order ablation.
The replay exposes a tradeoff between spatial admissibility and
query-conditioned relevance rather than establishing a universally optimal
ordering. In particular, within-Grounding Draft--Full agreement does not
measure the relevance excluded before Refinement. The main-paper
hierarchical-retrieval ablation (Table~3) instead measures downstream
Persistence and 3D Consistency for the complete retrieval design; those
generation scores and this proxy-recall diagnostic answer different
questions and do not establish a single causal geometric mechanism.

\begin{table}[t]
\centering
\normalsize
\caption{\textbf{Geometry--relevance trade-off on SolarWM-33B.}
Three trajectories and 72 repeated states with matched token budgets. Proxy recall is not spatial ground truth; reverse-order is a confounded diagnostic.}
\label{tab:app_geometry_draft_diagnostic_new}
\setlength{\tabcolsep}{3pt}
\begin{tabular}{lcc}
\toprule
Strategy & Proxy R@K $\uparrow$ & BFO=0 $\downarrow$ \\
\midrule
Draft-only & 93.91\% & 6.26\% \\
Geometry $\rightarrow$ Draft & 57.32\% & 5.27\% \\
Draft $\rightarrow$ Geom. (diag.) & 64.74\% & 0.18\% \\
\bottomrule
\end{tabular}
\end{table}

\subsection{Draft Refinement Fidelity}
\label{sec:appendix-refinement-fidelity}
\label{sec:appendix-retrieval-fidelity}

Draft Refinement approximates Full-Resolution Block Relevance on the same query state, scoring domain and block cap. In the separate five-case evaluation summarized in the main paper, Draft Refinement yields Pearson 0.989, Spearman 0.990, Kendall's $\tau$ 0.948, Recall@$K$ 95.62\%, NDCG@$K$ 0.9922, and Top-1 agreement 80.60\%. The observation-level Grounding Recall@4 (95.42\%) and MRR (0.740) are reported in Appendix~\ref{sec:appendix-grounding}. Table~\ref{tab:app_refinement_fidelity} adds archived cross-backbone cohorts; their results are reported separately and are not pooled with the five-case evaluation. Correlations and ranking metrics are summarized across trajectories, rather than treating all repeated states as independent samples.

\begin{table*}[t]
\centering
\caption{Cross-backbone Draft Block Relevance versus Full-Resolution Block Relevance in the Grounding domain, fixed $K=64$. Values are trajectory means $\pm$ sample SD; repeated layer--step--query states are not independent trajectories. The archived ORBIA cohorts are additional evidence, separate from the five-case aggregate in the main paper.}
\label{tab:app_refinement_fidelity}
\normalsize
\setlength{\tabcolsep}{1.4pt}
\renewcommand{\arraystretch}{1.10}
\begin{tabular}{lcccccccc}
\toprule
Backbone & \#Traj & \#States & Pearson & Spearman & Kendall & Recall@64 & NDCG@64 & Top-1 \\
\midrule
LingBot-1.3B & 5 & 150 & $0.991\,\mathord{\pm}\,0.002$ & $0.989\,\mathord{\pm}\,0.004$ & $0.945\,\mathord{\pm}\,0.007$ & $0.967\,\mathord{\pm}\,0.003$ & $0.996\,\mathord{\pm}\,0.001$ & $0.947\,\mathord{\pm}\,0.018$ \\
LingBot-14B & 5 & 150 & $0.982\,\mathord{\pm}\,0.004$ & $0.976\,\mathord{\pm}\,0.002$ & $0.906\,\mathord{\pm}\,0.004$ & $0.935\,\mathord{\pm}\,0.008$ & $0.994\,\mathord{\pm}\,0.002$ & $0.940\,\mathord{\pm}\,0.068$ \\
SolarWM-33B & 3 & 72 & $0.961\,\mathord{\pm}\,0.011$ & $0.971\,\mathord{\pm}\,0.007$ & $0.908\,\mathord{\pm}\,0.009$ & $0.914\,\mathord{\pm}\,0.007$ & $0.976\,\mathord{\pm}\,0.007$ & $0.750\,\mathord{\pm}\,0.150$ \\
\bottomrule
\end{tabular}
\end{table*}

\paragraph{The fixed max-branch weight.}
The production setting uses $\beta=0.04$ for both Draft and
Full-Resolution Block Relevance across all evaluated models and layers.
As established by Lemma~\ref{lem:pooling}, the mean branch is recovered
exactly under matched scoring and normalization domains, while the
approximation error is confined to the $\beta$-weighted max branch.
We report cross-backbone fidelity using the original archived cohorts in
Table~\ref{tab:app_refinement_fidelity}; no separate numerical
$\beta$-sweep is interpreted as an independently verified estimate of
that table's cohort. The following attention-output replay compares
mean-only and the fixed production weight under its explicitly stated,
limited evaluation supports, without implying that $\beta=0.04$ is
universally optimal.

\paragraph{Utility of the max branch.}
At the matched camera-domain retrieval budget, $\beta=0.04$ slightly lowers
relative $\ell_2$ output error from $0.040621$ to $0.039949$ over the
36 historical-only states and from $0.020845$ to $0.020623$ over the
13 verified resident-context states. Non-protected historical mass retained
changes from $87.171\%$ to $87.523\%$ and from $81.655\%$ to $82.151\%$,
respectively. In the secondary full-eligible-history domain (36 states),
output error decreases from $0.241015$ to $0.239367$, while mass retained
decreases from $42.931\%$ to $42.854\%$. Each mass fraction uses its stated
domain. The effects are small and depend on the evaluation domain and error
metric, so we do not interpret $\beta=0.04$ as a universally optimal fusion
weight.

\paragraph{Attention-output utility of retrieved blocks.}
Our primary output comparison uses 13 recorded states from one native
LingBot-1.3B trajectory with verified sink/current value tensors
(Table~\ref{tab:app_true_utility_new}). The replay preserves the shared
resident context and historical $\mathcal C_t\cup\mathcal P_t$ under
matched selection budgets. Draft and full-resolution
relevance yield nearly identical attention-output error in this verified
resident-context cohort: $0.020623$ versus $0.020552$ in relative $\ell_2$
error. The advantage over random selection is smaller once the shared
sink/current context is restored, because these resident tokens carry a
large fraction of the total attention mass. Only 13 recorded states contain
the complete verified sink/current value tensors required for this
comparison, so we treat it as a controlled replay rather than a broad
generation-level claim. The replay fixes native Q/K/V and positions; it
does not simulate selection-dependent temporal relocation.

\begin{table*}[t]
\centering
\normalsize
\caption{\textbf{True-attention utility with verified resident context.}
Means over 13 states from one LingBot-1.3B trajectory with shared resident
context and historical $(\mathcal C_t\cup\mathcal P_t)$ under matched
selection budgets. Historical mass is the retained fraction of
non-protected historical attention mass; full-context mass includes the
shared resident and protected tokens. Output error is
$\|O_{\mathrm{sel}}-O_{\mathrm{full}}\|_2/\|O_{\mathrm{full}}\|_2$
against the full-context reference. Native Q/K/V and positions are fixed.
Exact-mass and single-block sensitivity rankings are diagnostic selectors,
not joint-set output-error oracles.}
\label{tab:app_true_utility_new}
\begingroup
\setlength{\tabcolsep}{7pt}
\renewcommand{\arraystretch}{1.12}
\begin{tabular}{lccc}
\toprule
Selection & \shortstack{Non-protected history\\mass retained (\%) $\uparrow$}
& \shortstack{Full-context\\mass retained (\%) $\uparrow$}
& \shortstack{Output relative\\$\ell_2$ $\downarrow$} \\
\midrule
Random & 74.36 & 95.20 & 0.023822 \\
Uniform & 74.55 & 95.24 & 0.024139 \\
Recency & 69.48 & 94.49 & 0.031902 \\
Mean-only ($\beta=0$) & 81.66 & 96.28 & 0.020845 \\
Draft ($\beta=0.04$) & 82.15 & 96.38 & 0.020623 \\
Full relevance & 81.87 & 96.33 & 0.020552 \\
Exact-mass ranking & 82.24 & 96.42 & 0.025083 \\
Single-block sensitivity & 82.86 & 96.56 & 0.015686 \\
\bottomrule
\end{tabular}
\endgroup
\end{table*}

The secondary 36-state historical-only comparison has relative $\ell_2$
errors of $0.039949$ for Draft, $0.039875$ for Full relevance, $0.040621$
for mean-only, and $0.175155$ for Random. This comparison normalizes
attention over the historical camera-candidate domain and excludes
sink/current context; it therefore does not represent full-context
attention-output error. Separate native-QK cohorts provide correlations
with token-softmax block mass (Draft versus Full Spearman:
$0.491/0.462$ for LingBot-1.3B and $0.809/0.790$ for SolarWM-33B).

\begin{table}[t]
\centering
\caption{Full-Resolution relevance mass retained by Draft-selected blocks at $K=64$. Retention is $\sum_{b\in\widehat S_K}I_b/\sum_{b\in S_K^*}I_b$; regret is one minus retention. These are trajectory-balanced empirical diagnostics. Archived ORBIA and new WorldAtlas Solar cohorts are not merged.}
\label{tab:app_relevance_retention}
\normalsize
\setlength{\tabcolsep}{2pt}
\renewcommand{\arraystretch}{1.10}
\begin{tabular}{lcrr}
\toprule
Backbone / cohort & Traj/states & Retention & Regret \\
\midrule
LingBot-1.3B & 5/150 & 99.698\% & 0.302\% \\
LingBot-14B & 5/150 & 99.495\% & 0.505\% \\
SolarWM-33B (arch.) & 3/72 & 98.674\% & 1.326\% \\
SolarWM-33B (new) & 3/36 & 98.585\% & 1.415\% \\
\bottomrule
\end{tabular}
\end{table}

Relevance-mass retention in Table~\ref{tab:app_relevance_retention} complements set overlap: boundary substitutions can change the selected set while losing little Full-Resolution relevance. The new SolarWM WorldAtlas cohort is reported separately from archived ORBIA, since capture templates and trajectories differ. In a fixed-workload GPU-ready scorer diagnostic, increasing the historical archive from 16 to 81 latent frames raises scoring latency from 0.441 to 0.688\,ms, while Recall@64 remains between 95.94\% and 97.50\% against Exact-A2 selection.

\subsection{Spatial Granularity}
\label{sec:appendix-granularity}
\label{sec:appendix-spatial-granularity}

Table~\ref{tab:retrieval_granularity} provides the complete comparison of
chunk-, frame-, and block-level retrieval under the same active KV budget.
Block-level retrieval achieves the highest Recall@$K$ and relevance coverage
while distributing the retrieval budget across a broader set of historical
observations.

\begin{table}[!t]
\centering
\caption{
Ablation of retrieval granularity.
We compare different retrieval granularities against full-resolution block relevance.
}
\label{tab:retrieval_granularity}

\setlength{\tabcolsep}{2pt}
\renewcommand{\arraystretch}{1.12}
\normalsize

\begin{tabular}{lccc}
\toprule
\textbf{Granularity} &
\shortstack{\textbf{Recall@K}\\$\uparrow$} &
\shortstack{\textbf{Coverage}\\$\uparrow$} &
\shortstack{\textbf{Frames}\\\textbf{Touched}} \\
\midrule

Chunk Mean
& 74.86\% & 77.93\% & 8 \\

Chunk Max
& 72.15\% & 77.22\% & 8 \\

Frame Mean
& 88.87\% & 83.15\% & 8 \\

Frame Max
& 87.67\% & 82.85\% & 8 \\

\midrule
\textbf{Block}
& \textbf{98.01\%}
& \textbf{86.52\%}
& 11.31 \\

\bottomrule
\end{tabular}
\end{table}

To quantify the additive relevance opportunity directly, we additionally solve the optimal whole-frame selection under the same $K$-block cap. Every feasible whole-frame set is a feasible block set; the comparison therefore measures the relevance-allocation opportunity described by Remark~\ref{rem:granularity}, rather than generation quality.

\begin{table*}[t]
\centering
\caption{Block Top-$K$ versus the exact optimal whole-frame selection under the same block cap, using Full-Resolution Block Relevance. Relative gain is averaged over trajectories. Absolute gain is decomposed into equal-consumed-block gain and spare-slot gain: production $K$ need not be divisible by blocks per frame (16 for LingBot; 9 for SolarWM). Thus not all gain is attributed to within-frame heterogeneity.}
\label{tab:app_block_vs_frame}
\normalsize
\setlength{\tabcolsep}{3.2pt}
\renewcommand{\arraystretch}{1.10}
\begin{tabular}{llrcrr}
\toprule
Backbone & $K$ policy & Relative gain & Strict-gain states & Equal-consumed gain & Spare-slot gain \\
\midrule
LingBot-1.3B & 64 & 3.48\% & 146/150 & 0.010663 & 0.000000 \\
LingBot-1.3B & Production & 4.31\% & 149/150 & 0.008789 & 0.011975 \\
LingBot-14B & 64 & 10.52\% & 150/150 & 0.027031 & 0.000000 \\
LingBot-14B & Production & 9.44\% & 150/150 & 0.021042 & 0.023584 \\
SolarWM-33B & 64 & 6.47\% & 72/72 & 0.010257 & 0.002438 \\
SolarWM-33B & Production & 2.83\% & 63/72 & 0.007535 & 0.004245 \\
\bottomrule
\end{tabular}
\end{table*}

The whole-frame optimum uses actual block counts and an exact discrete selection, with no fractional frame. Table~\ref{tab:app_block_vs_frame} separates equal-consumed-block gains from gains due to spare slots, so indivisible whole-frame capacity is not misattributed to spatial heterogeneity. All 63 non-saturated SolarWM production states have strict gain, whereas nine saturated states, in which the budget admits every candidate block so both rules coincide, have zero gain, since both rules then select every candidate block. Relevance is nonuniformly distributed among Grounding candidates, motivating fine-grained block selection over uniform whole-frame retrieval. Across the matched SolarWM queries, actual production refinement touches 27.31 frames on average, with mean narrowing of 15.05\%. Finer allocation need not reduce the touched-frame count in every state, and score concentration does not establish pixel localization.

\subsection{Retrieval Controls and Diagnostics}
\label{sec:appendix-retrieval-controls}

We test whether position, recency and persistent historical-frame tendencies account for the descriptive peaks. Table~\ref{tab:app_retrieval_controls} keeps the archived scores unchanged. Its across-query comparison ranks the intersection of available history within each trajectory/layer/step, avoiding an advantage from newly available frames.

\begin{table}[t]
\centering
\caption{Combined retrieval controls (trajectory means). These are diagnostic retrieval controls, not independent generation-quality experiments. Across-query ranks use a shared-history intersection. Recency control excludes one complete native chunk immediately before the protected set, without renormalizing archived scores. BFO-positive coverage concerns the sampled estimator.}
\label{tab:app_retrieval_controls}
\normalsize
\setlength{\tabcolsep}{2pt}
\renewcommand{\arraystretch}{1.10}
\begin{tabular}{lrr}
\toprule
Control & 14B & 33B \\
\midrule
Initial/non-initial mean ratio & 1.003 & 1.458 \\
Maximum frame is chunk-initial & 12.00\% & 41.67\% \\
Own Top-1 rank on other queries & 13.50 & 8.18 \\
Across-query Top-1 persistence & 10.67\% & 21.43\% \\
Recency exclusion: zero-score blocks & 6.24\% & 8.72\% \\
Recency exclusion: zero-score mass & 6.48\% & 8.07\% \\
Eligible BFO-positive coverage by $C$ & 39.89\% & 73.16\% \\
\bottomrule
\end{tabular}
\end{table}

SolarWM shows a stronger chunk-initial association than LingBot-14B, but peaks are not exclusively chunk-initial. In the case0005 example, F15 remains rank 1 among chunk-initial frames in both displayed layers. Persistent frame tendencies exist across queries, but F15 is not permanently Top-1 (Top-1 frequencies 37.5\% and 25.0\% at layers 12 and 37). Fixed recency exclusion does not remove the sampled-zero-score effect. No causal video-VAE mechanism is inferred. The BFO-positive coverage row also shows that Top-$M$ does not retain every sampled-positive historical frame; sampled positivity is not an exact geometric intersection label.

\begin{table}[t]
\centering
\caption{Full-frame perceptual correspondence on six independent native LingBot-14B trajectories (72 fixed states). Spearman correlations are trajectory means $\pm$ sample SD. DINO uses cached DINOv1 ViT-B/16. LPIPS is raw distance; positive correlation therefore does not mean positive perceptual agreement. Both subsets retain the same original global-score normalization.}
\label{tab:app_perceptual}
\normalsize
\setlength{\tabcolsep}{0.8pt}
\renewcommand{\arraystretch}{1.10}
\begin{tabular}{llcc}
\toprule
Score & Competitive subset & DINO & Raw LPIPS \\
\midrule
Draft & Captured non-$P$ & $-0.043\,\mathord{\pm}\,0.140$ & $0.039\,\mathord{\pm}\,0.133$ \\
Draft & Production eligible & $-0.166\,\mathord{\pm}\,0.169$ & $0.131\,\mathord{\pm}\,0.126$ \\
Full & Captured non-$P$ & $-0.022\,\mathord{\pm}\,0.126$ & $0.023\,\mathord{\pm}\,0.125$ \\
Full & Production eligible & $-0.148\,\mathord{\pm}\,0.160$ & $0.117\,\mathord{\pm}\,0.118$ \\
\bottomrule
\end{tabular}
\end{table}

On six independent native LingBot-14B trajectories, Table~\ref{tab:app_perceptual} finds no broad positive full-frame association with DINO/LPIPS. Historical representatives are compared with the fixed current-frame representatives, using their mean similarity; current visual references were not produced by the tested remote retrieval selection. Cached DINOv1 ViT-B/16 is used, with full-frame resize and official LPIPS preprocessing. The separate local-region diagnostic yields DINO Spearman 0.466832 for Draft and 0.466386 for Full-Resolution relevance, with nearly identical local behavior. Local-region and full-frame metrics measure different correspondences. Draft Block Relevance preserves query-native feature relevance; neither score is a full-frame pixel-similarity metric or an approximation to PSNR. These independent full-frame diagnostics therefore do not validate a general pixel-level localization claim; the matched peaks in main-paper Fig.~5 remain a representative case.

\subsection{Grounding and Refinement Example}
\label{sec:appendix-retrieval-example}
\label{sec:appendix-case0005}
\label{sec:appendix-additional-visualizations}

Main-paper Fig.~5 shows a representative SolarWM-33B revisit in which
Draft relevance and frame-level PSNR peak at the same historical frame
in two layers. The example illustrates a possible correspondence within
the grounded candidates, not a general relationship between block-level
relevance and PSNR or a population-level localization result.

The generated-reference trajectory uses DraftKV retrieval, so the
example is descriptive rather than independent causal validation.
Table~\ref{tab:app_grounding_coverage} separately measures candidate
admission and within-candidate peak localization; these should not be
conflated with the case-specific alignment in main-paper Fig.~5.

\setcounter{topnumber}{5}
\setcounter{dbltopnumber}{4}
\setcounter{totalnumber}{9}
\renewcommand{\topfraction}{0.95}
\renewcommand{\dbltopfraction}{0.95}
\renewcommand{\textfraction}{0.03}
\renewcommand{\floatpagefraction}{0.75}
\renewcommand{\dblfloatpagefraction}{0.75}
\makeatletter
\setlength{\@fptop}{0pt}
\setlength{\@fpsep}{10pt}
\setlength{\@fpbot}{0pt plus 1fil}
\setlength{\@dblfptop}{0pt}
\setlength{\@dblfpsep}{10pt}
\setlength{\@dblfpbot}{0pt plus 1fil}
\makeatother

\section{Streaming and Systems Analysis}
\label{sec:appendix-systems}
\label{sec:appendix-memory-e2e}

\subsection{Streaming CG-Attention Ablations}
\label{sec:appendix-streaming-ablation}
\label{sec:appendix-streaming-efficiency}

Main-paper Fig.~7 summarizes fixed-query Fetch+Attn latency and active
KV reuse-buffer scaling. Table~\ref{tab:streaming_ablation} examines
matched selected-KV execution as the fetch size grows.
Tables~\ref{tab:app_camera_streaming_profile} and
\ref{tab:app_draft_streaming_profile} are separate fixed-query profiles
under camera-only and Draft-only retrieval, respectively. Their absolute
latencies and memory measurements are not repeats of main-paper Fig.~7,
and none of these fixed-query profiles is a 480-frame WorldAtlas
complete-generation E2E measurement. Comparisons of Serial versus
Streaming are made only within each matched selected-KV workload.

\begin{table*}[t]
\centering
\caption{
Latency and memory ablation of Streaming CG-Attention on an H200 GPU (SP1).
Both methods consume identical, preselected historical KV at different fetch sizes,
without retrieval scoring. Speedup is measured relative to serial execution.
}
\label{tab:streaming_ablation}

\normalsize
\setlength{\tabcolsep}{2.5pt}
\renewcommand{\arraystretch}{1.12}

\begin{tabular*}{\textwidth}{
@{\extracolsep{\fill}}
ll
cccc
cc
@{}
}
\toprule

\multirow{2}{*}{\textbf{Fetch Size}}
&
\multirow{2}{*}{\textbf{Method}}
&
\multicolumn{4}{c}{\textbf{Latency and Overlap}}
&
\multicolumn{2}{c}{\textbf{GPU Memory}}
\\

\cmidrule(lr){3-6}
\cmidrule(lr){7-8}

&
&
\textbf{Fetch+Attn}
&
\textbf{Speedup}
&
\textbf{H2D Active}
&
\textbf{Hidden H2D}
&
\textbf{Receive KV}
&
\textbf{Chunk Peak}
\\[-2pt]

&
&
\textbf{(ms) $\downarrow$}
&
\textbf{$\uparrow$}
&
\textbf{(ms) $\downarrow$}
&
\textbf{(\%) $\uparrow$}
&
\textbf{(MiB) $\downarrow$}
&
\textbf{(GiB) $\downarrow$}
\\

\midrule

\multirow{2}{*}{\shortstack{25 frames\\(689 MiB)}}
& Serial
& 34.597
& --
& 20.429
& 0.00
& 897.04
& 78.654
\\

& \textbf{Streaming}
& \textbf{17.451}
& \textbf{1.98$\times$}
& \textbf{12.925}
& \textbf{39.39}
& \textbf{32.00}
& \textbf{77.756}
\\

\midrule

\multirow{2}{*}{\shortstack{50 frames\\(1378 MiB)}}
& Serial
& 62.304
& --
& 40.546
& 0.00
& 1586.10
& 79.475
\\

& \textbf{Streaming}
& \textbf{30.196}
& \textbf{2.06$\times$}
& \textbf{25.703}
& \textbf{54.29}
& \textbf{32.00}
& \textbf{77.953}
\\

\midrule

\multirow{2}{*}{\shortstack{75 frames\\(2067 MiB)}}
& Serial
& 92.800
& --
& 60.883
& 0.00
& 2275.20
& 80.588
\\

& \textbf{Streaming}
& \textbf{42.868}
& \textbf{2.17$\times$}
& \textbf{38.535}
& \textbf{54.89}
& \textbf{32.00}
& \textbf{78.461}
\\

\bottomrule
\end{tabular*}

\end{table*}

\begin{table}[t]
\centering
\caption{
GPU staging-buffer memory under increasing
historical KV fetch budgets.
Prefetch buffering grows with the retrieval budget,
whereas Streaming CG-Attention maintains a
fixed 32 MiB receive buffer in this setting.
}
\label{tab:app_buffer_scaling}

\setlength{\tabcolsep}{2pt}
\renewcommand{\arraystretch}{1.12}
\normalsize

\begin{tabular}{cccc}
\toprule
\textbf{Frames} &
\shortstack{\textbf{Next-Layer}\\\textbf{Prefetch}\\\textbf{(MiB)}} &
\shortstack{\textbf{Next-Step}\\\textbf{2 Buffers}\\\textbf{(MiB)}} &
\shortstack{\textbf{Streaming}\\\textbf{(MiB)}} \\
\midrule

25 & 1794.08 & 2691.12 & \textbf{32} \\
50 & 3172.20 & 4758.30 & \textbf{32} \\
75 & 4550.33 & 6825.49 & \textbf{32} \\

\bottomrule
\end{tabular}
\end{table}

\begin{table*}[t]
\centering
\caption{
Detailed profiling of serial and Streaming CG-Attention
execution under camera-only retrieval (a separate fixed-query workload,
not a repeated main-paper Fig.~7 or WorldAtlas 480-frame E2E measurement).
Both methods consume identical selected historical KV.
Latency measures fetch-plus-attention execution;
GPU memory is reported for attention and chunk-level
NVML peaks, while Receive KV denotes the temporary
KV receive buffer.
}
\label{tab:app_camera_streaming_profile}

\normalsize
\setlength{\tabcolsep}{3pt}
\renewcommand{\arraystretch}{1.10}
\begin{tabular}{@{}llrrrrrrrr@{}}
\toprule
& & \multicolumn{5}{c}{\textbf{Latency and overlap}} & \multicolumn{3}{c}{\textbf{GPU memory}} \\
\cmidrule(lr){3-7}\cmidrule(lr){8-10}
\textbf{Backbone} & \textbf{Method} &
\shortstack{\textbf{Fetch}\\(MiB)} &
\shortstack{\textbf{Fetch}\textbf{+Attn}\\(ms) $\downarrow$} &
\shortstack{\textbf{Speedup}\\$\uparrow$} &
\shortstack{\textbf{H2D}\\(ms) $\downarrow$} &
\shortstack{\textbf{Hidden}\\\textbf{H2D}(\%) $\uparrow$} &
\shortstack{\textbf{Attn. Peak}\\(GiB) $\downarrow$} &
\shortstack{\textbf{Chunk Peak}\\(GiB) $\downarrow$} &
\shortstack{\textbf{Recv. KV}\\(MiB) $\downarrow$} \\
\midrule
\multirow{2}{*}{LingBot 1.3B} & Serial & \multirow{2}{*}{132.539} & 6.239 & 1.000$\times$ & 3.420 & 0.00 & 5.450 & 8.754 & 167.883 \\
& \textbf{Streaming} & & \textbf{4.062} & \textbf{1.536$\times$} & \textbf{2.495} & \textbf{62.44} & \textbf{5.337} & \textbf{8.619} & \textbf{48} \\
\midrule
\multirow{2}{*}{LingBot 14B} & Serial & \multirow{2}{*}{441.797} & 20.563 & 1.000$\times$ & 11.990 & 0.00 & 39.819 & 43.451 & 559.609 \\
& \textbf{Streaming} & & \textbf{12.287} & \textbf{1.674$\times$} & \textbf{8.344} & \textbf{36.69} & \textbf{39.326} & \textbf{42.965} & \textbf{40} \\
\midrule
\multirow{2}{*}{SolarWM 33B} & Serial & \multirow{2}{*}{964.688} & 46.175 & 1.000$\times$ & 28.445 & 0.00 & 69.624 & 74.090 & 1175.590 \\
& \textbf{Streaming} & & \textbf{21.923} & \textbf{2.106$\times$} & \textbf{18.024} & \textbf{47.91} & \textbf{68.524} & \textbf{72.938} & \textbf{32} \\
\bottomrule
\end{tabular}
\end{table*}

\begin{table*}[t]
\centering
\caption{
Detailed profiling of serial and Streaming
CG-Attention execution under Draft-only retrieval
(without Draft Grounding). This is a separate fixed-query workload,
not a repeated main-paper Fig.~7 or WorldAtlas 480-frame E2E measurement.
Both methods consume identical selected historical KV.
Latency measures fetch-plus-attention execution;
GPU memory is reported for attention and chunk-level
NVML peaks, while Receive KV denotes the temporary
KV receive buffer.
}
\label{tab:app_draft_streaming_profile}

\normalsize
\setlength{\tabcolsep}{3pt}
\renewcommand{\arraystretch}{1.10}
\begin{tabular}{@{}llrrrrrrrr@{}}
\toprule
& & \multicolumn{5}{c}{\textbf{Latency and overlap}} & \multicolumn{3}{c}{\textbf{GPU memory}} \\
\cmidrule(lr){3-7}\cmidrule(lr){8-10}
\textbf{Backbone} & \textbf{Method} &
\shortstack{\textbf{Fetch}\\(MiB)} &
\shortstack{\textbf{Fetch}\textbf{+Attn}\\(ms) $\downarrow$} &
\shortstack{\textbf{Speedup}\\$\uparrow$} &
\shortstack{\textbf{H2D}\\(ms) $\downarrow$} &
\shortstack{\textbf{Hidden}\\\textbf{H2D}(\%) $\uparrow$} &
\shortstack{\textbf{Attn. Peak}\\(GiB) $\downarrow$} &
\shortstack{\textbf{Chunk Peak}\\(GiB) $\downarrow$} &
\shortstack{\textbf{Recv. KV}\\(MiB) $\downarrow$} \\
\midrule
\multirow{2}{*}{LingBot 1.3B} & Serial & \multirow{2}{*}{123.703} & 5.951 & 1.000$\times$ & 3.256 & 0.00 & 5.653 & 8.508 & 159.047 \\
& \textbf{Streaming} & & \textbf{3.896} & \textbf{1.527$\times$} & \textbf{2.325} & 3.65 & \textbf{5.545} & \textbf{8.371} & \textbf{48} \\
\midrule
\multirow{2}{*}{LingBot 14B} & Serial & \multirow{2}{*}{412.344} & 19.952 & 1.000$\times$ & 11.748 & 0.00 & 40.607 & 43.686 & 530.156 \\
& \textbf{Streaming} & & \textbf{11.802} & \textbf{1.691$\times$} & \textbf{7.791} & \textbf{37.89} & \textbf{40.128} & \textbf{43.033} & \textbf{40} \\
\midrule
\multirow{2}{*}{SolarWM 33B} & Serial & \multirow{2}{*}{689.063} & 35.628 & 1.000$\times$ & 20.532 & 0.00 & 70.370 & 74.109 & 899.965 \\
& \textbf{Streaming} & & \textbf{20.169} & \textbf{1.766$\times$} & \textbf{12.957} & \textbf{36.86} & \textbf{69.523} & \textbf{73.229} & \textbf{32} \\
\bottomrule
\end{tabular}
\end{table*}

\paragraph{Runtime slot-count sensitivity.}
We vary the receive-slot count on one H200 using fixed deterministic
BF16 synthetic Q/K/V, selected tokens,
positions and attention partitions through the native FA3 execution path.
Actual receive storage increases from 2 to 8\,MiB, while median warmed
Fetch+Attention time is 5.80--5.90\,ms in this fixture. All outputs are
bitwise equal to the fixed-partition native dense reference; independent
math-SDPA attention on the same positioned KV also passes the stated
numerical tolerance. These repeated fixture calls do not establish a
full-model optimum. Single-slot Streaming remains distinct from Serial
selected-set materialization.

\subsection{Bandwidth and Prefetch Analysis}
\label{sec:appendix-bandwidth}

The oracle lookahead strategies are diagnostic references, not production baselines, and Streaming is not universally fastest across bandwidth settings.

\begin{table*}[t]
\centering
\caption{
Bandwidth sensitivity under different historical KV
fetch budgets. We compare Next-Step Oracle, Prefetch,
and Streaming CG-Attention.
Service rates are measured in GB/s, while next-step
latencies are reported in seconds.
Unlimited denotes no imposed bandwidth cap.
}
\label{tab:app_bandwidth_sensitivity}

\setlength{\tabcolsep}{5pt}
\renewcommand{\arraystretch}{1.12}
\normalsize

\begin{tabular}{ccccccc}
\toprule
\textbf{Frames} &
\textbf{Cap (GB/s)} &
\multicolumn{2}{c}{\textbf{Service Rate (GB/s)}} &
\multicolumn{3}{c}{\textbf{Next-Step Latency (s) $\downarrow$}} \\
\cmidrule(lr){3-4}
\cmidrule(lr){5-7}
& &
\textbf{Prefetch} &
\textbf{Streaming} &
\textbf{Oracle} &
\textbf{Prefetch} &
\textbf{Streaming} \\
\midrule

25 & 10.215 & 10.16 & 10.18
& 3.542 & 3.580 & 5.398 \\

25 & 12.769 & 12.68 & 12.71
& 2.881 & 3.000 & 4.690 \\

25 & 15.961 & 15.81 & 15.87
& 2.860 & 3.000 & 4.128 \\

25 & Unlimited & 35.22 & 55.09
& 2.844 & 2.903 & \textbf{2.738} \\

\midrule

50 & 20.430 & 20.27 & 20.37
& 3.542 & 3.613 & 5.053 \\

50 & 25.538 & 25.31 & 25.42
& 2.916 & 3.066 & 4.343 \\

50 & 31.922 & 31.54 & 29.70
& 2.910 & 3.055 & 3.952 \\

50 & Unlimited & 35.23 & 55.71
& \textbf{2.915} & 3.419 & 2.934 \\

\midrule

75 & 30.646 & 30.34 & 29.25
& 3.702 & 3.877 & 4.781 \\

75 & 38.307 & 35.39 & 37.86
& 3.268 & 3.735 & 4.214 \\

75 & 47.883 & 35.32 & 40.01
& 3.139 & 3.827 & \textbf{3.777} \\

75 & Unlimited & 35.23 & 55.78
& 3.132 & 4.166 & \textbf{3.079} \\

\bottomrule
\end{tabular}
\end{table*}

Oracle prefetch requires sufficient transfer bandwidth and lookahead
capacity to approach zero gating. For larger retrieval budgets, near-zero
gating is not always reached within the tested bandwidth range; the
observed threshold can also be non-monotonic.

\subsection{Persistent Memory Scaling}
\label{sec:appendix-memory-scaling}
\label{sec:appendix-draft-bank-scaling}

The Draft K Bank stores compact descriptors for accumulated historical latent frames. Its logical payload and the CPU full-resolution KV archive both grow linearly:
\begin{equation}
 M_{\mathrm{DraftBank}}(N)=\Theta(N),\qquad
 M_{\mathrm{CPU\text{-}KV}}(N)=\Theta(N).
\end{equation}
Only the active KV Reuse Buffer has $O(1)$ capacity with respect to history/retrieved-context length for fixed $(R,C,G,d,b)$, as specified in Appendix~\ref{sec:appendix-streaming}. This is not a bound on total GPU memory.

\begin{table}[!htbp]
\centering
\caption{Persistent Draft K Bank descriptor-payload scaling. $N$ counts historical latent frames; per-rank slopes include all layers. The bank is compact but grows linearly with accumulated history. Only the active KV Reuse Buffer is bounded with respect to retrieved-context length at fixed configuration. Ratios use native full BF16 K+V; allocator, pointer-directory and raw temporal-key auxiliary bytes are separate.}
\label{tab:app_draft_bank_scaling}
\normalsize
\setlength{\tabcolsep}{0.6pt}
\renewcommand{\arraystretch}{1.10}
\begin{tabular}{lccccc}
\toprule
Backbone & SP & \shortstack{MiB/frame\\per rank} & \shortstack{GiB/1k\\frames\\all ranks} & \shortstack{Bank/full\\K+V} & $R^2$ \\
\midrule
LingBot-1.3B & 1 & 2.812 & 2.747 & 1.061\% & 1.000 \\
LingBot-14B & 8 & 1.562 & 12.207 & 1.061\% & 1.000 \\
SolarWM-33B & 8 & 1.538 & 12.016 & 0.893\% & 1.000 \\
\bottomrule
\end{tabular}
\end{table}

Capacity probes at $N=16,32,64,128,256,512$ fit $M(N)=aN+c$
with $R^2=1$, $c=0$ bytes and maximum payload residual 0 for all
three backbones (Table~\ref{tab:app_draft_bank_scaling}). The logical
descriptor payload occupies 0.89--1.06\% of full BF16 K+V; these are
storage probes rather than generated long trajectories.

\paragraph{Host-archive accounting and memory scope.}
Table~\ref{tab:app_memory_scope_new} reports logical GPU descriptor-bank
counters, separately from host allocation and process RSS. Allocator
reservation, pointer directories and raw temporal-key auxiliaries are
additional costs. Historical SP8 per-rank counters and SP1 measurements
have different accounting scopes.

\begin{table}[!htbp]
\centering
\normalsize
\caption{\textbf{Logical Draft K Bank counters for audited H200/SP1 runs.}
Values are not host pinned allocation, host RSS, or comparable-length GPU peaks.}
\label{tab:app_memory_scope_new}
\setlength{\tabcolsep}{5pt}
\begin{tabular}{lc}
\toprule
Backbone & Logical Draft K Bank (GiB) \\
\midrule
LingBot-1.3B & 0.253 \\
LingBot-14B & 1.123 \\
SolarWM-33B & 1.322 \\
\bottomrule
\end{tabular}
\end{table}

\paragraph{Sampled process host memory.}
Sampling rank-0 RSS at nominal two-second intervals over the approximate
generation window gives maxima of
182.72--182.95\,GiB for Serial and 182.74--182.95\,GiB
for Streaming across three LingBot-14B pairs. The SolarWM-33B pair gives
179.26 and 179.29\,GiB, respectively. RSS includes model/runtime and mapped
memory; these sampled maxima are neither archive-only allocation nor exact
peaks. Summed process RSS double-counts shared mappings. A separate
single-trajectory SolarWM-33B replay records 166.55\,GiB of successfully
CUDA-registered canonical-archive backing capacity. This component measure
excludes other pinned allocators and in-flight registrations. Total process
pinned allocation remains unmeasured; logical KV bytes and VmPin are not substitutes.

\paragraph{Full-model long-history execution.}
Table~\ref{tab:appendix-long-history-execution} measures one 120\,s
LingBot-1.3B continuation with all 30 layers and 12 heads on one H200.
It generates 1,920 decoded frames at 16\,fps and accumulates new KV history;
the camera-control prefix is extended by holding its last pose.
Across the recorded frontiers, full-model forward medians range from
789.645 to 888.701\,ms, while recorded live GPU allocation grows from
6.782 to 8.173\,GiB. Registered archive capacity and process RSS are
reported separately from GPU allocation. The archive counter includes
successfully registered, ready prefetch capacity and excludes other pinned
allocators. The full instrumented run includes initial compilation;
these raw within-run observations are not clean E2E timings. Query content
and the eligible history evolve together, so the table describes execution
along one generated trajectory rather than isolating a causal effect of
history length.

\subsection{Single-H100 Deployment Capability}
\label{sec:appendix-h100}

SolarWM-33B with DraftKV completes 357 frames on one physical H100 80GB
GPU under SP1, with an NVML-sampled peak of approximately 74.0\,GiB.
This 357-frame ORBIA-length deployment test is not a reproduction of
the main-paper Table~2 WorldAtlas 480-frame system measurement.
Its timing protocol differs from the controlled H200 comparisons, so this
run establishes deployment capability only. WorldKV also completes under
its own tested configuration; single-H100 feasibility is therefore not
exclusive to DraftKV. Native encounters OOM only in the tested configuration.

\subsection{End-to-End Streaming Validation}
\label{sec:appendix-streaming-e2e}
\label{sec:appendix-e2e-streaming}

H200 comparisons fix generation inputs, selected historical KV,
protected/sink/current routes, positions, temporal K-RoPE and attention
partitions while varying Serial materialization versus Streaming.
The complete-generation comparisons in this subsection use 357-frame
ORBIA cases, unlike the 30\,s, 480-frame WorldAtlas cases underlying
main-paper Table~2's E2E/FPS. The reported speedups compare Serial and
Streaming within each identical case; their absolute seconds are not
directly comparable with Table~2. The fixed-query Fetch+Attn tests
reported separately below also have a different measurement scope.
Measurements retain their recorded timer and warm-cache policies;
in-timer compiler activity excludes a run from latency statistics.

\begin{table*}[!htbp]
\centering
\caption{H200 repeatability. Panel A: three paired 357-frame ORBIA complete generations per backbone, reported as mean $\pm$ sample SD in seconds; speedup is the mean of paired ratios. These are not 480-frame WorldAtlas runs from main-paper Table~2. Panel B: within-process fixed-query Fetch+Attn, 10 warmups and 30 alternating pairs, in milliseconds. The 30-pair measurements are not 30 independent full generations. Complete-generation repeats span dates and H200 cards and use one fixed case per backbone.}
\label{tab:app_e2e_repeated}
\normalsize
\setlength{\tabcolsep}{2.2pt}
\renewcommand{\arraystretch}{1.10}
{\normalsize\textbf{A. Complete generation (s)}}\par\smallskip
\begin{tabular}{lcccc}
\toprule
Backbone & $n$ & Serial (s) & Streaming (s) & Speedup \\
\midrule
LingBot-14B & 3 & $327.855\,\mathord{\pm}\,8.951$ & $301.544\,\mathord{\pm}\,12.189$ & 1.088$\times$ \\
SolarWM-33B & 3 & $616.566\,\mathord{\pm}\,8.882$ & $535.641\,\mathord{\pm}\,13.003$ & 1.151$\times$ \\
\bottomrule
\end{tabular}
\par\medskip
{\normalsize\textbf{B. Fixed-query Fetch+Attn (ms)}}\par\smallskip
\begin{tabular}{lcccc}
\toprule
Backbone & Warmups & Pairs & Serial (ms) & Streaming (ms) \\
\midrule
LingBot-14B & 10 & 30 & $19.354\,\mathord{\pm}\,0.053$ & $12.758\,\mathord{\pm}\,0.053$ \\
SolarWM-33B & 10 & 30 & $36.017\,\mathord{\pm}\,0.126$ & $22.244\,\mathord{\pm}\,0.722$ \\
\bottomrule
\end{tabular}
\end{table*}

Panel A of Table~\ref{tab:app_e2e_repeated} repeats one fixed profiling
case per backbone across dates and same-model H200 cards, using fresh
processes and matched warm-cache policies. LingBot includes conditioning
and VAE decoding but excludes video encoding; SolarWM uses verified cached
conditioning and includes video writing. Panel B measures a separate
within-process fixed query; its repeats do not count as independent generations.

Separate native diagnostic traces measure H2D active/non-overlapped-with-attention activity of 35.217/26.451\,s for LingBot-14B and 60.618/42.710\,s for SolarWM-33B, corresponding to about 24.9\% and 29.5\% overlap. Scoring plus Top-$K$ host scopes account for 15.96\% and 1.87\% of the respective instrumented Streaming generation durations. These instrumented, potentially nested scopes are not an additive causal decomposition or the clean repeated E2E timings.

Streaming reduces exposed historical-KV access cost by removing the
selected-set materialization barrier. Scoring, model compute, non-overlapped
transfer and pipeline fill/drain remain; H2D overlap alone does not explain
the complete-generation gain.

\paragraph{Trajectory-level backend controls.}
Table~\ref{tab:appendix-multitrajectory-e2e} adds three distinct LingBot-14B
trajectories and two SolarWM-33B trajectories under the same paired controls.
Ordered selection/source-map records and complete videos agree within each
pair. Both backends include the same verification observer, whose incremental
overhead is unmeasured. These 357-frame matched-observer backend controls are not
remeasurements of the 480-frame WorldAtlas latency in main-paper Table~2.

\begin{table}[!htbp]
\centering
\caption{\textbf{Additional H200/SP1 trajectory-level backend controls.}
Each row is one 357-frame complete-generation pair using BF16 and TP1.
S and St denote Serial and Streaming; speedup is S/St.
Runs use warm compilation caches and retain first-use runtime initialization.
Compiler-contaminated and resource-flagged observations are excluded.
LingBot-14B has three admitted trajectories; SolarWM-33B has two.}
\label{tab:appendix-multitrajectory-e2e}
\normalsize
\setlength{\tabcolsep}{3pt}
\renewcommand{\arraystretch}{1.10}
\begin{tabular}{lrrr}
\toprule
Backbone / case & S (s) & St (s) & Speedup \\
\midrule
LingBot-14B / 0005 & 323.38 & 313.56 & 1.031$\times$ \\
LingBot-14B / 0013 & 330.68 & 292.07 & 1.132$\times$ \\
LingBot-14B / 0051 & 323.03 & 288.97 & 1.118$\times$ \\
SolarWM-33B / 0005 & 599.07 & 542.79 & 1.104$\times$ \\
SolarWM-33B / 0051 & 608.61 & 544.91 & 1.117$\times$ \\
\bottomrule
\end{tabular}
\end{table}

Across the three LingBot-14B trajectories, the paired speedup has mean
$1.094\times$, sample SD $0.055$ and median
$1.118\times$. Across the two SolarWM-33B trajectories, the paired
speedup has mean and median $1.110\times$ and sample SD $0.009$.
The models use separate physical H200 GPUs with target-GPU process and
shared-host monitoring. Their timer scopes follow those specified above,
so raw seconds are not unconditional cross-backbone latency comparisons.

\begin{table}[!htbp]
\centering
\caption{\textbf{Full-model long-history execution diagnostic.}
One generated LingBot-1.3B trajectory on H200/SP1/TP1, BF16.
History counts archived latent frames, not decoded frames or retrieval budgets.
Panel A reports host-wall medians over four denoising forwards at each
frontier; Fetch+Attention sums native calls across all layers.
These instrumented scopes are inclusive and must not be added together;
the full run includes compilation and is not a clean E2E benchmark.
Panel B samples ready registered archive capacity at the last denoising
forward and takes RSS/live PyTorch GPU-allocation maxima over the five
forward-complete boundaries of that chunk. Registered capacity includes
ready prefetch backing and is not total process pinned memory.}
\label{tab:appendix-long-history-execution}
\normalsize
\setlength{\tabcolsep}{3pt}
\renewcommand{\arraystretch}{1.10}
\begin{tabular}{rrrr}
\toprule
\multicolumn{4}{c}{Panel A: within-run latency (ms)} \\
\midrule
History & \shortstack{Scorer\\+ TopK} & \shortstack{Fetch\\+ Attention} & \shortstack{Full-model\\forward} \\
\midrule
32 & 31.622 & 122.349 & 799.804 \\
64 & 31.711 & 122.060 & 789.645 \\
128 & 31.371 & 123.390 & 815.940 \\
256 & 31.358 & 122.447 & 844.661 \\
384 & 31.902 & 123.256 & 874.669 \\
480 & 31.928 & 127.760 & 888.701 \\
\midrule
\multicolumn{4}{c}{Panel B: recorded memory (GiB)} \\
\midrule
History & \shortstack{Registered\\archive} & RSS & \shortstack{Live GPU\\allocated} \\
\midrule
32 & 14.561 & 38.202 & 6.782 \\
64 & 24.269 & 47.933 & 6.881 \\
128 & 43.684 & 67.365 & 7.080 \\
256 & 82.513 & 106.250 & 7.477 \\
384 & 121.343 & 145.121 & 7.875 \\
480 & 146.825 & 170.614 & 8.173 \\
\bottomrule
\end{tabular}
\end{table}

\end{document}